\documentclass[referee]{sn-jnl}

\jyear{2021}%

\usepackage[T1]{fontenc}
\usepackage[utf8]{inputenc}
\usepackage[english]{babel}

\theoremstyle{thmstyleone}%
\theoremstyle{thmstyletwo}%
\usepackage{graphicx}
\graphicspath{{Figures/}}
\usepackage[square,numbers]{natbib}
\usepackage{multirow}
\usepackage{enumerate}
\usepackage{booktabs}
\usepackage{amsmath,amssymb}
\usepackage{array}
\usepackage[normalem]{ulem}
\usepackage{comment}
\usepackage{amsfonts}
\usepackage{amsthm}
\usepackage{mathtools, cuted}
\usepackage{xr-hyper}
\usepackage{hyperref}
\usepackage{thmtools}

\theoremstyle{thmstylethree}%
\newtheorem{definition}{Definition}%

\declaretheoremstyle[
spaceabove=6pt, spacebelow=6pt,
headfont=\normalfont\bfseries,
notefont=\mdseries, notebraces={(}{)},
bodyfont=\normalfont,
postheadspace=0.6em,
headpunct=:
]{mystyle}
\declaretheorem[style=mystyle, name=Hypothesis, preheadhook={}]{hyp}

\usepackage{cleveref}
\crefname{hyp}{hypothesis}{hypotheses}
\Crefname{hyp}{Hypothesis}{Hypotheses}

\algnewcommand\algorithmicforeach{\textbf{for each}}
\algdef{S}[FOR]{ForEach}[1]{\algorithmicforeach\ #1\ \algorithmicdo}

\newcommand{\J}[1]{\textcolor{blue}{#1}}

\newcommand{\PP}[1]{\textcolor{purple}{#1}}

\newcommand{\G}[1]{\textcolor{brown}{#1}}

\makeatletter
\newcommand*{\addFileDependency}[1]{
  \typeout{(#1)}
  \@addtofilelist{#1}
  \IfFileExists{#1}{}{\typeout{No file #1.}}
}
\makeatother

\newcommand*{\myexternaldocument}[1]{%
    \externaldocument{#1}%
    \addFileDependency{#1.tex}%
    \addFileDependency{#1.aux}%
}

\begin{document}

\title[Efficient Task Allocation in Smart Warehouses]{Efficient Task Allocation in Smart Warehouses with Multi-delivery Stations and Heterogeneous Robots}


\author*[1]{\fnm{George} \sur{S.~Oliveira}}\email{george.oliveira@posgrad.ufsc.br}

\author[2]{\fnm{Juha} \sur{Röning}}\email{juha.roning@oulu.fi}
\equalcont{These authors contributed equally to this work.}

\author[1]{\fnm{Patricia} \sur{D.~M.~Plentz}}\email{patricia.plentz@ufsc.br}
\equalcont{These authors contributed equally to this work.}

\author[1]{\fnm{Jônata} \sur{T.~Carvalho}}\email{jonata.tyska@ufsc.br}
\equalcont{These authors contributed equally to this work.}

\affil*[1]{\orgdiv{Department of Informatics and Statistics (INE)}, \orgname{Federal University of Santa Catarina}, \orgaddress{\street{Campus Universitário Reitor João David Ferreira Lima, Trindade}, \city{Florianópolis}, \postcode{88040-900}, \state{Santa Catarina}, \country{Brazil}}}

\affil*[2]{\orgdiv{Biomimetics and Intelligent Systems Group - BISG, Finland}, \orgname{University of Oulu},
\orgaddress{\city{Oulu}, \postcode{}, \country{Finland}}}


\abstract{The task allocation problem in multi-robot systems (MRTA) is an NP-hard problem whose viable solutions are usually found by heuristic algorithms. Considering the increasing need of improvement on logistics, the use of robots for increasing the efficiency of logistics warehouses is becoming a requirement. In a smart warehouse the main tasks consist of employing a fleet of automated picking and mobile robots that coordinate by picking up items from a set of orders from the shelves and dropping them at the delivery stations. Two aspects generally justify multi-robot task allocation complexity: (i) environmental aspects, such as multi-delivery stations and dispersed robots (since they remain in constant motion) and (ii) fleet heterogeneity, where robots' traffic speed and capacity loads are different from each other. Despite these properties have been widely researched in the literature, they usually are investigated separately. Also, many algorithms are not scalable for problems with thousands of tasks and hundreds of robots. This work proposes a scalable and efficient task allocation algorithm for smart warehouses with multi-delivery stations and heterogeneous fleets. Our strategy employs a novel cost estimator, which computes costs as a function of the robots' variable characteristics and capacity while they receive new tasks. For validating the strategy a series of experiments is performed simulating the operation of smart warehouses with multiple delivery stations and heteregenous fleets. The results show that our strategy generates routes costing up to $33$\% less than the routes generated by a state-of-the-art task allocation algorithm and $96$\% faster in test instances representing our target scenario. Considering single-delivery stations and non-dispersed robots, we reduced the number of robots by up to $18$\%, allocating tasks $92$\% faster, and generating routes whose costs are statistically similar to the routes generated by the state-of-the-art algorithm.}




\maketitle

\section{Introduction}
\label{sec:introduction}

E-commerce is a type of commercial transaction carried out through online platforms where consumers purchase products and services. This type of transaction keeps growing and has become more and more critical over the years. In 2020, Online sales volume expressively increased due to SARS-CoV-2 pandemic, forcing the logistics and transportation sectors to improve their product delivery methods.

In this process of buying and selling products online a key component is the logistics warehouse. The online orders are automatically sent to these warehouses, and they are responsible for grouping products, packing, and shipping orders. Due to the ever increasing volume of online transactions, manually operating logistics warehouses becomes unfeasible, which requires the development and use of automation technologies for warehouses.

According to \citet{Berg1999519}, a set of equipment and policies applied to manage goods in warehouses defines a \emph{warehouse system}. The automation level of a warehouse system categorizes such a system as manual, automated, robotic, or smart \cite{DACOSTABARROS2021103729}. According to \citet{Berg1999519}, and \citet{DACOSTABARROS2021103729}, a smart warehouse comprises the employment of computational intelligence techniques, together with Internet of Things (IoT) technologies, to carry out analytical approaches that employ teams of robots for carrying out order pickings. The main purposes of this approach is reducing human intervention and optimizing warehouse operations.

According to \citet{Bolu202127246}, a \emph{Robotic Mobile Fulfillment System} is used to designate a smart warehouse that uses mobile robots in their operations. The robots lift \emph{pods} and take them from \emph{picking stations} to \emph{delivery stations} for manual order sorting. The first RMFS-based warehouses used the KIVA system (current Amazon Robotics) \cite{Wurman20081, Boysen20181}. However, once Amazon's robot fleet became exclusive to the company, many others entered the market, such as inVia Robotics\copyright, LOCUS Robotics\copyright, OMRON Automation \& Safety\copyright, OTTO Motors\ copyright, Fetch Robotics\copyright, Hitachi\copyright, among others.

As described in \cite{Conveyco_guide}, the main reason to employ an RMFS is to provide greater security, accuracy, productivity, and efficiency in smart warehouses. A RMFS employs multi-robot systems (MRS) that achieve a common goal that would be impossible using a single robot. However, robots need to work coordinately seeking to optimize processes and avoid conflicts to achieve the expected efficiency. One of the important challenge for MRS research is designing appropriate coordination strategies between robots that allow them to perform operations efficiently in terms of time and workspace \cite{Yan20131}. The reason for the growing interest in multi-robot systems is the anticipated benefits when compared to single-robot systems (SRS), such as solving complex tasks, increasing overall performance, and system reliability \cite{Khamis201531}.

Within the research on MRS, multi-robot task allocation (MRTA) is one of the most studied problems. The MRTA aims to ensure that the total cost of performing all tasks is minimized while minimizing the number of robots \cite{Sarkar20185022}. Another approach is to meet different constraints and complete one or more general objectives, which usually includes performing all tasks \cite{Wang20091} and minimizing the makespan \cite{Dou20151, Xue2019TaskAO}, traveled distance \cite {Sarkar20185022}, or energy consumption \cite{Pinkam2016269, Li201617}.

An efficient solution needs to consider constraints that can impact execution costs or increase problem complexity. Such constraints may include relocatable tasks (\cite{GOMBOLAY2018220, Yinbin2021776}), deadline (\cite{Li201617, Xue2019TaskAO, Edelkamp2019288, Sarkar20191052}), environment's uncertainties  (\cite{ Claes2017492, Edelkamp2019288}) and different robots' properties such as load capacity (\cite{Sarkar20185022, Xue2019TaskAO}), traffic speed (\cite{Dou20151, Yinbin2021776}) and energy consumption (\cite{Pinkam2016269, Li201617, Xue2019TaskAO}). In smart warehouses, constraints still include the dispersion of robots, location and demand of the tasks, number of delivery stations, warehouse layout, and other environmental features such as obstacles, roadblocks, and other unpredictable situations.

The main tasks performed in smart warehouses consist on collecting goods contained in a set of orders from the shelves and drop them off to the delivery stations.
The main objective is to fulfill all orders upholding constraints and minimizing the objective function's cost.
The objective functions include, for example, (1) reducing the time to fulfill orders, (2) reducing the robots' energy consumption, and (3) reducing the number of robots used to perform all tasks.


\subsection{Motivation}
\label{subsec:motivation}

Regarding smart warehouses, there is trend that automated goods/products gathering using picking robots (\cite{IEEE_Spectrum_Fetch_Robotics, Winkelhaus20211}) will keep increasing. Furthermore, according to \citet{Parker2008}, the scope of solvable tasks of MRS increases with the use of heterogeneous robots, allowing parallelism and robustness, which leads to better performance with complex and straightforward robot teams. Therefore, considering smart warehouse scenarios, heterogeneous fleets of robots are useful. This type of fleet offers economic benefits as it becomes cheaper to distribute varied tasks among robots with different capabilities and costs, instead of many costly robots with the same properties \cite{Parker2008}.

By investigating state-of-the-art works from the perspective of constraints in smart warehouses, we observed a gap in exploring the possibility of using heterogeneous robot fleets in smart warehouses. For example, we found works that address the MRTA where robots have different load capacities from each other (\cite{Claes2017492}), but as far as we could find out, there are no works that explore robots with different load capacities and traffic speeds at the same time. It is noteworthy, such distinction is common even between robots from the same manufacturer.

In addition to these constraints, environmental aspects such as the warehouse layout, initial (standby) positions of the robots, number and positions of delivery stations, directly influence warehouse performance. In general, warehouses have multiple product delivery stations and due to the high demand for orders, robots are constantly dispersed across the map in constant motion. For the sake of optimization in logistics, we believe that it is interesting that robots remain strategically dispersed, even if in standby, to optimize the picking and delivery of goods. Although works about MRTA consider such factors individually, we observe a lack of studies that address all the constraints mentioned above.

Therefore, assuming that heterogeneous fleets will be constant in smart warehouses with automated pickup, there is a need to develop task allocation strategies that consider multiple constraints and different objective function components, such as time, traveled distance, and energy consumption. We hypothesize that formulating the problem and the objective function, taking into account a cost estimator measured as a function of all such constraints, allows us to generate better cost estimates and, consequently, more efficient task assignments. We briefly present the problem below.


\subsection{Problem Description}
\label{subsec:problem}

In a seminal work developed by \citet{Gerkey2004939}, an MRTA problem is classified according to its taxonomy: 

\begin{enumerate}
	\item \textbf{Multi-Task Robots (MT) vs. Single-Task Robots (ST)}, whether multiple tasks can be assigned to the same robot or not;
	\item \textbf{Multi-Robot Tasks (MR) vs. Single-Robot Tasks (SR)} whether the tasks require robots to form coalitions or not;
	\item \textbf{Time-extended Assignment (TA) vs. Instantaneous Assignment (IA)}, if new tasks appear while robots perform others.
\end{enumerate}

Overall, an MRTA problem of any kind is computationally intractable for large-scale applications \cite{Gerkey2004939, Nunes2017ATF}. For this reason, most solution methods are approximate, i.e., they do not guarantee that solutions will have the minimum overall cost, but they are fast and provide solutions that are acceptable in practice. In this context, there are works that employ auction-based algorithms \cite{LEE2018151}, heuristics \cite{Sarkar20185022, Xue2019TaskAO, Sarkar20191052}, meta-heuristics \cite{Li2020AMF, Tsang20181671, Li201617} and hybrid algorithms \cite{Claes2017492, Pinkam2016269, Dou20151}, for example.

In this work, we propose a mechanism for a Robotic Mobile Fulfillment System that considers many constraints. Robots can receive multiple tasks simultaneously; each task only needs one robot to be finished, and there is no need to reallocate tasks. The warehouse employs automated pickup, has several delivery stations, and keeps robots dispersed to promote optimization. Robots can visit delivery stations as often as necessary to fulfill all tasks. In the taxonomy of \citet{Gerkey2004939}, the proposed solution seeks to solve an MT-SR-IA problem.

Works on smart warehouses sometimes treat MT-SR-IA as one of the variations of the classic vehicle routing problem (VRP) \cite{Christofides197610}. VRP was introduced by \citet{Dantzig195980} as a generalization of the traveling salesman problem (TSP) to deal with the problem of transporting items by multiple vehicles. Such a problem is well known in operations research where one or more depots supply a group of known demand and location customers. The objective is to determine one or more routes through which a fleet of vehicles must depart to serve customers within a minimum distance.

Over the years, the VRP has had several variations, with different objectives and constraints determined by the properties of the problem. For example, in the capacitated vehicle routing problem (CVRP), vehicles have limited load capacity, and the objective is to determine routes so that vehicles serve customers without the sum of demands exceeding their load capacity \cite{Sarkar20185022, Edelkamp2019288, Kloetzer20191579}. In the multi-depot vehicle routing problem (MDVRP), in addition to the previous features, vehicles depart from different depots \cite{Cordeau1997105}. The heterogeneous fleet vehicle routing problem (HFVRP) is similar to CVRP but considers a heterogeneous fleet of vehicles, generally differing in load capacity, and individual cost \cite{Panicker20181}. Vehicles start and end in the same depots in all of these variations.

In this work, we approach MT-SR-IA from the perspective of the vehicle routing problem with heterogeneous fleet, many depots, and dispersed vehicles (HFMDVRP-DV).  In other words, the robot fleet (vehicles) is heterogeneous and dispersed on the map, and our scenario owns several stations for delivering products (multi-depot). Many other works approached VRP to address the multi-robot task allocation in warehouses \cite{Claes2017492, Sarkar20185022, Edelkamp2019288, Sarkar20191052, Kloetzer20191579, Rubrico200858}, and other indoor environments \cite{Parlaktuna2007187, Khalifa2011726}. However, to the best of our knowledge, this is the first work that addresses the HFMDVRP-DV, and such VRP variation applied indoors. The assumptions for this work are:

\begin{itemize}
	\item Robots are initially dispersed on the map.
	\item Robots must collect goods with different demands (weight) spread across the warehouse;
	\item Robots collect goods according to their load capacities;
	\item There are several delivery stations where robots have to deliver goods. Delivery stations are visited when (i) robots need to restore their load capacity, or (ii) after collecting the last good, to ensure delivery of all of them;
	\item The robot fleet is heterogeneous, i.e., the robots are different in their load capacity and traffic speed;
	\item Robots deliver products to any delivery station;
	\item Robots visit delivery stations as many times as necessary to ensure that the capacity constraint is not violated.
\end{itemize}


\subsection{Objectives and Contributions}
\label{subsec:objectives_contrib}

In this work, we present a space decomposition-based heuristic to solve the HFMDVRP-DV called \textbf{Do}main zo\textbf{Ne} based \textbf{C}pacity and \textbf{P}riority constrained \ textbf{T}ask \textbf{A}llocator (DoNe-CPTA). DoNe-CPTA employs a cost estimator that takes into account differences in load capacity to determine the probability of robots visiting a delivery station. Our proposal manages to postpone visits to stations, generating lower-cost routes and increasing the robots' efficiency. DoNe-CPTA uses our cost estimator with an adaptation of the Voronoi diagram (\cite{Pokojski2018141}) to create the least costly task groups for each robot, namely \emph{domain zones}. We compared DoNe-CPTA with an adaptation of a state-of-the-art algorithm for smart warehouses called nCAR \cite{Sarkar20185022}, which was adjusted to meet the properties of the HFMDVRP-DV. Well-known benchmark datasets used for CVRP problems~\cite{UCHOA2017845} were used in their classical and adapted form. Results showed that DoNe-CPTA is efficient in execution time and execution cost, while minimizing the number of robots used. Thus, our main contributions are as follow:

\begin{enumerate}
	\item Presenting the mathematical formulation of the HFMDVRP-DV;
	\item Presenting a cost estimator that includes multiple factors and constraints;
	\item Developing an efficient task allocation algorithm for building efficient routes in fast time while minimizing the number of robots used;
	\item Since no other work has covered HFMDVRP-DV before, we have developed and introduced new datasets adapted from well-known instances of operations research datasets \cite{UCHOA2017845}. We use real robot specifications so that our datasets are close to real-world scenarios.
	\item Presenting results that indicate that DoNe-CPTA surpasses state-of-the-art solutions in terms of cost, robots, and execution time.
\end{enumerate}

Results show that DoNe-CPTA can generate routes costing up to $33$\% less using up to $18$\% fewer robots than the state-of-the-art algorithm.

This article is organized as follows: In Section \ref{sec:related}, we present the works related to smart warehouses and MRTA from the perspective of the previously presented constraints. Section \ref{sec:problem_formulation} presents the mathematical formulation of HFMDVRP-DV. Section \ref{sec:cost_estimator} presents our cost estimator model. Section \ref{sec:algorithm} presents the design of our decomposition-based heuristic DoNe-CPTA. We present the HFMDVRP-DV datasets in Section \ref{sec:datasets}. The experimental setup, the description of the state-of-the-art algorithm, and the validation of execution cost, time, and robot usage aspects we present in Section \ref{sec:evaluation}. Finally, Section \ref{sec:conclusion} brings the conclusions we draw based on the results obtained.
\section{Related Works}
\label{sec:related}

This section presents the related works from the perspective of all constraints that we enumerated in Section \ref{subsec:objectives_contrib}. Several works take into account the load capacity \cite{Pinkam2016269, Xue2019TaskAO, Claes2017492, Sarkar20185022, Edelkamp2019288, Kloetzer20191579} and some others deal with the traffic speed \cite{Dou20151, Xue2019TaskAO, Yinbin2021776, Edelkamp2019288}. These constraints are approached simultaneously in \cite{Xue2019TaskAO} and \cite{Edelkamp2019288}. \citet{Pinkam2016269}, \citet{Dou20151}, \citet{Li201617}, \citet{Yinbin2021776} and \citet{Kloetzer20191579} tackle the MRTA with robots starting dispersed on the map. \citet{Pinkam2016269}, \citet{Dou20151}, \citet{Xue2019TaskAO}, and \citet{Kloetzer20191579} attend scenarios with multiple delivery stations. Table \ref{tab:related_works} summarizes these works. We point out the particularities of each of these works below.

\begin{table*}[ht]
\begin{center}
\footnotesize
\begin{minipage}{\textwidth}
\caption{Related works from the perspective of the constraints mentioned in Section \ref{sec:introduction}. The last line refers to this work.}
\label{tab:related_works}
\begin{tabular*}{\textwidth}{@{\extracolsep{\fill}}cccccc@{\extracolsep{\fill}}}
\toprule
Papers & \begin{tabular}[c]{@{}c@{}}Load\\ Capacity\end{tabular} & \begin{tabular}[c]{@{}c@{}}Traffic\\ Speed\end{tabular} & \begin{tabular}[c]{@{}c@{}}Heterogeneous\\ Fleet\end{tabular} & \begin{tabular}[c]{@{}c@{}}Dispersed\\ Robots\end{tabular} & \begin{tabular}[c]{@{}c@{}}Multi-Delivery\\ Station\end{tabular} \\
\midrule
\citet{Pinkam2016269}                                                  & $\surd$                                                                &                                                                 &                                                                        & $\surd$                                                                & $\surd$                                                                        \\
\citet{Dou20151}                                                       &                                                                     & $\surd$                                                            &                                                                        & $\surd$                                                                & $\surd$                                                                        \\
\citet{Li201617}                                                       &                                                                     &                                                                 &                                                                        & $\surd$                                                                &                                                                             \\
\citet{Xue2019TaskAO}                                                  & $\surd$                                                                & $\surd$                                                            & $\star$                                                                      &                                                                     & $\surd$                                                                        \\
\citet{Yinbin2021776}                                                  &                                                                     & $\surd$                                                            &                                                                        & $\surd$                                                                &                                                                             \\
\citet{Claes2017492}                                                   & $\surd$                                                                &                                                                 &                                                                        &                                                                     &                                                                             \\
\citet{Sarkar20185022}                                                 & $\surd$                                                                &                                                                 &                                                                        &                                                                     &                                                                             \\
\citet{Edelkamp2019288}                                                & $\surd$                                                                & $\surd$                                                            & $\star$                                                                      &                                                                     &                                                                             \\
\citet{Sarkar20191052}                                                 &                                                                     &                                                                 &                                                                        &                                                                     &                                                                             \\
\citet{Kloetzer20191579}                                               & $\surd$                                                                &                                                                 &                                                                        & $\surd$                                                                & $\star$                                                                           \\
Our work                                                       & $\surd$                                                                & $\surd$                                                            & $\surd$                                                                   & $\surd$                                                                & $\surd$                                                                        \\
\botrule
\end{tabular*}
\end{minipage}
\end{center}
\end{table*}

We noticed that most of the works deal with robots' load capacity either in isolation or along with another constraint. Five works cover three constraints, and just one covers four constraints simultaneously. \citet{Xue2019TaskAO} and \citet{Claes2017492} study robots that differ either in load capacity or traffic speed, respectively. Also, although the theoretical approach of \citet{Edelkamp2019288} assumes heterogeneous robots, experiments only employ homogeneous robots. In \citet{Kloetzer20191579}, robots must deliver products to respective stations, but the warehouse contains more than one delivery station. We developed this work to deal with such constraints simultaneously, even though the results show that our strategy can also meet configurations without all these characteristics. The common thread is that all the works described in this section approach smart warehouses. In Table \ref{tab:related_works}, the last six works (including ours) deal with the MRTA as a reduction to one variation of VRP. We detail these works below into two groups: (i) MRTA strategies focused exclusively on smart warehouses, and (ii) MRTA strategies addressed as VRP problems (and their variations) that share characteristics of smart warehouse applications.

\subsection{MRTA in Smart Warehouses}

\citet{Pinkam2016269} assumes a warehouse where each picking station receives an order and waits for robots to bring the corresponding goods.
The proposed method allocate tasks with a greedy algorithm,  respecting the robots' load capacity. Then, robots use Dijkstra's algorithm to find the shortest path. Authors present two versions of the developed solution: the Individual Collection (IC), in which each robot is assigned to each picking station, and the Collaborative Collection (CC), in which all robots can pick up goods from all stations. Experiments compared the IC with CC measuring energy consumption and tasks' execution time. Results showed that the CC reduces the makespan by up to 15\% while keeping the same energy consumption. Like our proposal, this problem considers a warehouse with several delivery stations and allows robots to start from any place, but all robots and tasks must have the same load capacity and demand, respectively.

\citet{Dou20151} developed a genetic algorithm-based heuristic without coping with tasks' demands and robots' load capacities \cite{Holland199266}. Robots receive their tasks and employ reinforcement learning \cite{Arulkumaran201726} to avoid collisions during navigation and compute their rewards using distance and traffic speed. Robots always choose the action with the highest reward. Experiments compare this algorithm with other three task allocators: (i) auction-based, (ii) K-means-based \cite{Macqueen1967281} and (iii) random. For each allocation method, the robots plan routes in two different ways: by reinforcement learning and by $A$* \cite{Hart1968100}. The presented results indicated better performances of the genetic algorithm. Concerning path planning, $A$* showed better results than reinforcement learning, but it is suitable only when the scenario is known to all robots. With reinforcement learning, robots acquire the ability to adapt to different circumstances.

\citet{Li201617} presents a solution for MRTA in a warehouse with several picking stations. A central control unit receives orders and allocates them to the secondary control units associated with each delivery station. Then, secondary units process orders and allocate tasks to robots. Costs are computed in the function of distance, robots' energy cost (fixed for all robots), and tasks' deadline. Finally, robots must fulfill tasks respecting deadlines but not their capacity. Like our work, this study also considers the dispersion of robots throughout the warehouse. This work introduced an Improved Ant Colony Optimization (ACO) to adapt to the proposed cost estimator. Experiments showed better results compared to the original ACO \cite{Dorigo19921}.

\citet{Xue2019TaskAO} studied a warehouse where tasks are associated with demand, deadline, and an unmet penalty. Their method allocates tasks according to robots' load capacities and an energy consumption-based estimator. First, tasks are assigned to picking stations according to a special set of rules. After that, the energy consumption-based estimator computes costs based on distance, robots' traffic speed, tasks' deadlines, and tasks' unmet penalties. Robots have equal speeds and are initially dispersed on the map. After that, the secondary control units (underlying each warehouse picking station) assign tasks to robots. Experiments used the representation of an existing warehouse and the previous version of the proposed algorithm. Results showed improvements in task execution time and reduction in robots' energy consumption in all provided instances.

\citet{Yinbin2021776} deal with a warehouse with several shelves and some picking positions for each shelf. Robots are initially positioned in several parking areas scattered around the warehouse periphery. Their solution consists of a decentralized approach where the robots themselves provide the estimated time to complete the tasks. Robots compute their costs by taking into account the fleet speed (the fleet is homogeneous) but not the load capacity. A density prediction algorithm determines the density of robots scattered on the map, predicting their route, and determining possible conflicts. Then, the cost estimator adjusts the costs if the density prediction detects potential conflicts. The proposed method assigns the task to the robot with the lowest estimated cost. Finally, robots compute their routes with the Floyd-Warshall algorithm \cite{Floyd1962, Warshall196211, Ingerman1962550}. Experiments evaluated the efficiency of the density prediction in a simulated warehouse with $100$ shelves and up to $100$ robots. New tasks appear every 2 seconds up to a total of $200$ tasks. Results showed that the average time to complete all tasks is shorter when density prediction is active. The results also showed that increasing the number of robots increases the time to assign tasks but decreases the time for robots to fulfill them.

\subsection{VRP Variations in Smart Warehouses}
\label{subsec:vrp_var_smart_warehouse}

In the context of vehicle routing (VRP), \citet{Claes2017492} considers the probability of task execution failure to address real-world uncertainties. This problem can be reduced to the capacitated vehicle routing (CVRP) since the modeling involves a homogeneous fleet (same capacity for all robots) starting at a single point and one delivery station. Furthermore, there is the possibility of new tasks arising while robots perform others. Such a problem was solved using Monte Carlo Tree Search (MCTS) \cite{Coulom20061}, which is an exhaustive search simulation-based algorithm that tests samples from hundreds of possible trajectories to obtain good approximations for the values of possible actions. These samples consider the probability of failures caused by real-world uncertainties and the possibility of new tasks arising. After that, robots plan their routes using one of the three versions of the greedy algorithm: the simplest, where the robots address the tasks according to the lowest computed cost; the reverse greedy algorithm, where the method adapts depending on the change in the robots' position, and the iterative greedy algorithm, which iteratively evaluates the position of all robots to reallocate tasks. Experiments considered a simulated warehouse where new tasks could arise, and robots could fail. The introduced method was integrated with three versions of the greedy algorithm and then compared with the state-of-the-art EFWD \cite{Claes2015881} algorithm. Results show that the MCTS got the best rewards in all setups.

The article \cite{Sarkar20185022} presents nCAR (Nearest-neighbor based Clustering and Routing), a heuristic for task allocation in smart warehouses for solving the CVRP. First, nCAR creates feasible routes through a greedy search so that the tasks' demand does not exceed the vehicles' capacity. After that, each feasible route is transformed into an effective route using the Christofides' algorithm \cite{Christofides19761}. Experiments use two reference datasets for VRP problems: \textsc{P} \cite{Augerat19951} and \textsc{X} \cite{UCHOA2017845}. First, the proposed solution was compared with the state-of-art OR-Tools \cite{GoogleORTools2021} and optimal costs of the first dataset, showing that the method is able to generate sub-optimal routes quickly. After that, comparing nCAR only with OR-Tools when executing instances of dataset \textsc{X}, nCAR proved to be more efficient than the state-of-the-art algorithm, generating routes with less cost, less time, and with fewer robots.

\citet{Edelkamp2019288} introduced and studied the \emph{physical vehicle routing problem} (PVRP) as a way to address MRTA under constraints such as load capacity and traffic speed in smart warehouses. The fleet is heterogeneous, and robots are required to perform tasks within a specific deadline. Their scenario has a single point in common where all robots must depart, return and deliver the picked items. A variation of Nested Monte-Carlo Search \cite{Cazenave2009456}, called NMCS-based nested rollout policy adaptation (NRPA) \cite{Rosin2011649}, performs advanced searches in the search tree with reinforcement learning to weight actions and prioritize the movements of the better sequence of each step of NRPA and thus solve the PVRP. Experiments used instances of the \citet{Solomon1987254} dataset, which is widely used in research on VRP with deadlines and on varied environment and algorithm parameters, such as vehicles and the NRPA recursion level. Results showed that the NRPA produces routes with minor violations (in load capacity and tasks' deadline), at the same time that it significantly increases the computation time when the number of robots increases.

\citet{Sarkar20191052} tackled the MRTA in smart warehouses, where tasks are also associated with deadlines and unmet penalties. Robots must transport goods spread across the picking station to the (single) delivery station. The objective is to allocate tasks to robots so that the accumulated penalty is minimal without taking into account the robots' load capacities and traffic speeds. This work developed a two-step algorithm called Minimum Penalty Scheduling (MPS). First, the proposed algorithm assigns the maximum number of highest priority tasks, not violating deadlines, and then the MPS allocates the other tasks (lower priority tasks) to reduce the total penalty. Experiments used $12$ variations on instances with $100$ and $1000$ tasks of the dataset \textsc{X} to validate the MPS against the Auction-based Distributed Algorithm (ADA) \cite{Luo2015876}. Using a combination of scaling techniques adapted from the work of \citet{Cooper20021} together with load balancing techniques, MPS found solutions with significantly low penalties compared to ADA.

In \cite{Kloetzer20191579} the MRTA was reduced to the Capacity and Distance constrained Vehicle Routing Problem (CDVRP), which is a VRP variation where vehicles (robots) have load capacity and distance limits. Such a problem consists of transporting goods from picking to delivery stations in a warehouse using \emph{micro aerial vehicles} (drones). Each drone is assigned to a storage station and can only carry goods of the same type within their respective station. They depart from their station, pick up goods according to their capacity and deliver them to the same place from which they departed. This work modeled the problem's constraints and objectives in a binary integer programming (BIP) \cite{ChinneckBIP2021}, being optimally solved by CPLEX \cite{Laborie2018210}.

\subsection{Discussion}

Although these works also focus on legitimate aspects of real environments, such as uncertainties, time, and energy consumption, we tried to synthesize them from the perspective of other constraints, such as the heterogeneous fleet, the robots' dispersion, and multi-delivery stations. We noticed that only \cite{Pinkam2016269}, \cite{Xue2019TaskAO}, \cite{Claes2017492}, \cite{Sarkar20185022}, \cite{Edelkamp2019288} and \cite{Kloetzer20191579} take into account the load capacity. Although some properties of a heterogeneous fleet are present in \cite{Xue2019TaskAO} and \cite{Claes2017492}, the respective works did not take into account robots with different capacities and speeds simultaneously. We also noticed that some works~\cite{Claes2017492, Edelkamp2019288, Kloetzer20191579} are not scalable for problems with thousands of tasks, being unfeasible to meet real warehouse scenarios; and other works~\cite{Pinkam2016269, Xue2019TaskAO, Yinbin2021776} have not been validated by comparison to state-of-the-art algorithms.

\section{Problem Formulation}
\label{sec:problem_formulation}

\begin{figure*}[!htbp]
	\centering
	\includegraphics[width=1\textwidth]{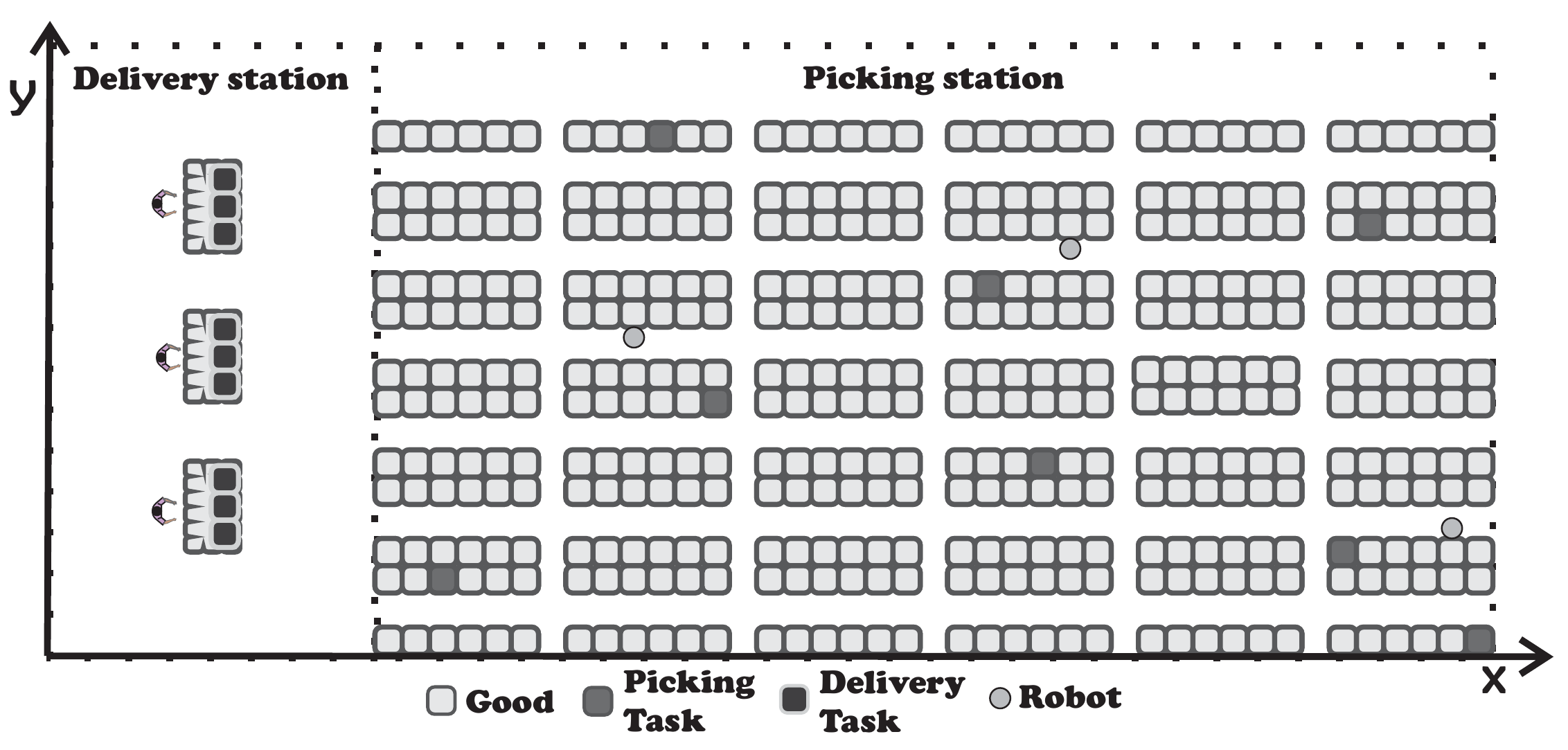}
	\caption{Example of warehouse scenario represented in a Cartesian plane.}
	\label{fig:warehouse}
\end{figure*}

This section presents the formulation of HFMDTA-DR for smart warehouses. The warehouse is mapped onto a Cartesian plane as shown in Figure \ref{fig:warehouse}. There are two types of stations: the picking station, containing shelved goods and aisles where robots can travel; and delivery stations, containing delivery points for goods. Every good is a potential task, which means that any goods lying around at the picking station can be picked up and transported to one of the delivery stations. All tasks have a non-zero \emph{demand}, i.e. weight, that must be satisfied by robots.

The robot fleet is heterogeneous, with each robot having a different traffic speed and load capacity. Robots are initially dispersed throughout the warehouse, and all of them can pick any product, i.e., can perform any task, as long as the sum of the tasks' demand does not exceed their load capacity.
There are tasks that are more demanding than the load capacity of some robots, but there will be at least one robot with enough capacity to perform the most demanding task at a given time.

This work deals with two types of tasks: (1) \emph{picking task}, where robots pick up items from shelves, and (2) \emph{delivery task}, where robots take the picked items to a delivery station. Robots perform delivery tasks when (i) they do not have the load capacity to perform the next picking task, or (ii) there are no other picking tasks. This ensures that all goods will be delivered to the delivery stations. Delivery tasks comprise visiting one of the delivery stations present in the warehouse.

In the following subsections, we formulate the MRTA problem for dealing with dispersed robots (vehicles), heterogeneous fleet, and multi-delivery stations (multi-depot). For the sake of clarity, we present the HFMDVRP-DV formulation by incrementally inserting such constraints into the CVRP formulation (\cite{Sarkar20185022}), which is our baseline formulation. Subsection \ref{subsec:baseline_formulation} presents both CVRP and MDVRP (\cite{Cordeau1997105}), the latter being a CVRP extension after dealing with multi-depots. Next, subsection \ref{subsec:baseline_formulation_heterogeneous}, adds heterogeneous fleet properties (\cite{Panicker20181}) in MDVRP. Finally, subsection \ref{subsec:baseline_formulation_heterogeneous_dispersed} presents our contribution on HFMDVRP-DV after considering dispersed robots. All formulations follow the respective VRP definition according to works on operational research \cite{Sarkar20185022, Cordeau1997105, Panicker20181}.

\subsection{Baseline Formulation and Multi-Delivery Stations}
\label{subsec:baseline_formulation}

Let $R = \{r_1, \ldots, r_n\}$ be the set of robots, $T = \{t_1, \ldots, t_m\}$ the set of picking tasks and $H = \{h_1, \ldots , h_p\}$ the set of delivery tasks. $G = (V, E)$ is a simple, undirected weighted graph with $m+p$ vertices. The set $V = V^T \cup V^H$ denotes the vertices of $G$, where $V^T$ is the set of vertices that correspond to the picking tasks and $V^H$ is the set of vertices that correspond to the delivery tasks (i.e., locations delivery stations). The starting position $u_0^i$ of each robot $r_i$ is the same as any delivery station. In other words, $u_0^i \in V^H$. $G$ has edges between:

\begin{itemize}
	\item each pair of vertices in $V^T$;
	\item between each vertex in $V^T$ with each vertex in $V^H$;
\end{itemize}

\noindent
so robots can start from the starting point ($u_0^i$), visit any picking task, and return to the starting point. There are no edges between pairs in $V^H$.

Each edge $e$ in $E$ has a cost $k^e$ for any robot $r_i \in R$ to travel through $e$. Each vertex $v_i \in V^T$ has a demand $d_i$ that must be satisfied. Let $C$ be the load capacity of all robots in $R$, the demand of any vertex cannot exceed the maximum capacity of the robots, i.e.:

\begin{align}
	d_i \leq C \enspace \forall \enspace v_i \in V^ T.
\end{align}

The objective of MDVRP is to find a set of cycles with disjoint vertices for every $r_i$, each robot starting at $u_0^i$. Each robot can have zero to $t$ cycles associated to it, as long as the cost of all cycles is minimized and the total demand on the vertices in each cycle does not exceed the robots' capacity. Each cycle is a route: 

\begin{align}
	A_i = (u_0^i, v_k, \ldots, u^i), v_k\in V^T \wedge u_0^i \in V^H
\end{align}

\noindent
whose cost $K_i$ is given by:

\begin{align}
	K_i(A_i) = k^{u_0^i \rightarrow v_1} + \ldots + k^{v_k \rightarrow u_0^i}
\end{align}

\noindent
where $u \rightarrow v$ is the edge in $E$ connecting the vertices $u$ and $v$ and $k^{u \rightarrow v}$ is the cost of the edge $u \rightarrow v$.

Thus, the ideal MDVRP solution is to find $t$ routes $A_i^1, A_i^2, \ldots, A_i^t$, for each robot $r_i$ where (1) the sum of all routes is minimized; (2) all routes of $r_i$ are determined by vertices in $V^T$, except $u_0^i$; (3) the only common point in each route of $r_i$ is $u_0^i$ and (4) the demand of all tasks in each route does not exceed the maximum capacity $C$ of robots. Mathematically:

\begin{description}
	\item $\sum_{i=1}^{n} \sum_{j=1}^{t} K_i(A_i^j)$ is minimized;
	\item $\bigcup_{j=1}^{t} A_i^j \enspace \backslash \enspace \{v_0^i\} = V^T, \enspace 1 \leq i \leq n$;
	\item $\bigcap_{j=1}^{t} A_i^j = \{v_0^i\}, \enspace 1 \leq i \leq n$;
	\item $\sum_{v_k \in A_i^j} d_k \leq C, \enspace 1 \leq i \leq n$;
\end{description}

CVRP is an instance of the MDVRP with only one delivery station, i.e., $H = \{h_1\}$ and $V^H = \{v_0\}$. All robots start from $v_0$ and must return to $v_0$ at the end of the route.

\subsection{Heterogeneous Fleet}
\label{subsec:baseline_formulation_heterogeneous}

In Heterogeneous Fleet and Multi-Delivery Task Allocation (HFMDTA), the MDTA definitions need to be adjusted to include the individual properties of each robot. In the case of this work, such property refers to traffic speed. Thus, it is necessary to define a graph $G^i$ separately for each robot $r_i$. The cost of each edge $e \in E^i$ is defined as a function of one or more properties of $r_i$.

Furthermore, the robots' load capacities and traffic speed are distinct, as the fleet is heterogeneous. Let $C^\ominus$ be the smallest load capacity among all robots. Then, the demand of any vertex cannot exceed the smallest capacity among robots, i.e., $d_i \leq C^\ominus \enspace \forall \enspace v_i \in V^T$.

The objective of HFMDTA is to find a set of cycles with disjoint vertices for each robot $r_i$, with each robot starting at its respective $u_0^i$. Each robot $r_i$ can have zero to $t$ cycles associated with it, as long as the cost of all cycles is minimized and the total demand of the vertices in each cycle does not exceed $C_i$. Each cycle is a route: 

\begin{align}
	A_i^j = (u_0^i, v_k, \ldots, u_0^i), v_k \in V^T \wedge u_0^i \in V^H
\end{align}

\noindent
whose cost $K_i$ is given by:

\begin{align}
	K_i(A_i^j) = k_i^{u_0^i \rightarrow v_1} + \ldots + k_i^ {v_k \rightarrow u_0^i}
\end{align}

\noindent
where:
\begin{enumerate}
	\item $A_i^j$ is the j-th route of $r_i$;
	\item $u \rightarrow v$ is the edge in $E^i$ between the vertices $u$ and $v$;
	\item $k_i^{u \rightarrow v}$ is the cost of the edge $u \rightarrow v$ measured as a function of the robot's traffic speed;
	\item $K_i(A_i^j)$ is the cost of the j-th route of $r_i$ measured as a function of the $r_i$'s traffic speed.
\end{enumerate}

Mathematically, the HFMDTA's ideal solution is to find $t$ routes $A_i^1, A_i^2, \ldots, A_i^t$, for each robot $r_i$ such that:

\begin{description}
	\item $\sum_{i=1}^{n} \sum_{j=1}^{t} K_i(A_i^j)$ is minimized;
	\item $\bigcup_{j=1}^{t} A_i^j \enspace \backslash \enspace \{v_0^i\} = V^T, \enspace 1 \leq i \leq n$;
	\item $\bigcap_{j=1}^{t} A_i^j = \{v_0^i\}, \enspace 1 \leq i \leq n$;
	\item $\bigcap_{i=1}^{n} \bigcap_{j=1}^{t} A_i^j \subseteq V^H$;
	\item $\sum_{v_k \in A_i^j} d_k \leq C_i, \enspace 1 \leq i \leq n$;
\end{description}

\subsection{Dispersed Robots}
\label{subsec:baseline_formulation_heterogeneous_dispersed}

Unlike previous variations, in HFMDVRP-DV the robots' starting point is not necessarily the location of any delivery station but any location in the map. $G^i = (V^i, E^i)$ is a simple, undirected weighted graph with $m+p+1$ vertices associated to each robot $r_i$. $v_0^i$ is the vertex that corresponds to the starting position of $r_i$, which does not necessarily belong to $V^H$. $G$ has edges between:

\begin{description}
	\item each pair of vertices in $V^T$;
	\item between $v_0$ and each vertex in $V^T$;
	\item between the vertices of $V^H$ and $V^T$;
\end{description}

\noindent
so that $r_i$ can start from its starting point $v_0^i$, visit any picking task, and go back to any delivery station in $V^H$ to perform a delivery task. There are no edges between $v_0$ and any vertex in $V^H$.

Each edge $e \in E^i$ is costed as a function of the $r_i$'s traffic speed.

As robots start from varying positions, the objective of the HFMDVRP-DV is to find a set of paths (not cycles) for each $r_i$, with each robot starting at its respective $v_0^i$. The last vertex of each path is a vertex in $V^H$. Mathematically, each path is a route:

\begin{align}
	A_i^j = (v_0^i, \alpha_{i:2}, \ldots, \beta), \alpha_{i:k} \in V^T \wedge \beta \in V^H
\end{align}

\noindent
whose cost $K_i$ is given by:

\begin{align}
	K_i(A_i^j) = k_i^{v_0^i \rightarrow \alpha_{i:2}} + k_i^{\alpha_{i:2} \rightarrow \alpha_{i:3}} + \ldots + k_i^ {\alpha_{i:a-1} \rightarrow \beta}
\end{align}

Therefore, the ideal solution of the HFMDVRP-DV is to find $t$ routes $A_i^1, A_i^2, \ldots, A_i^t$, for each robot $r_i$ such that:

\begin{description}
	\item $\sum_{i=1}^{n} \sum_{j=1}^{t} K_i(A_i^j)$ is minimized;
	\item $\bigcup_{j=1}^{t} A_i^j \enspace \backslash \enspace  V^T \subseteq \{v_0^i\} \cup V^H, \enspace 1 \leq i \leq n$;
	\item $\bigcup_{j=1}^{t} A_i^j \enspace \backslash \enspace  V^H \subseteq \{v_0^i\} \cup V^T, \enspace 1 \leq i \leq n$;
	\item $\bigcap_{j=1}^{t} A_i^j \enspace \backslash \enspace \{v_0^i\} \subseteq V^H \enspace 1 \leq i \leq n$;
	\item $\bigcap_{i=1}^{n} \bigcap_{j=1}^{t} A_i^j \subseteq V_0 \cup V^H \enspace \mid \enspace V_0 = v_0^1 + \ldots + v_0^n$;
	\item $\sum_{v_k \in A_i^j} d_k \leq C_i, \enspace 1 \leq i \leq n$;
\end{description}
\section{Cost Estimator Based on Variable Load Capacity}
\label{sec:cost_estimator}

In the previous section, we presented the HFMDTA-DR formulation without explaining how the costs are estimated. This section introduces a cost estimator that considers the distance between robots and tasks, traffic speed, and robots' maximum and variable load capacity. Costs are first defined by the distance between robot $r$ and task $t$ and the traffic speed of $r$. So, when comparing the variable load capacity of $r$ and the demand of $t$, if $r$ needs to visit a delivery station before or after executing $t$, then the previously estimated cost

is incremented with the cost of performing a delivery task. The basic premise is to penalize robots with low variable load capacity, giving way to other robots and thus postponing visits to delivery stations as much as possible.

Let $\Gamma_i$ be the maximum capacity of the robot $r_i$ and $\gamma_i^j$ the capacity of $r_i$ after executing the picking task $t_j$. We say that a picking task $t_k$ is \emph{absolutely feasible} to $r_i$ (or $r_i$ is absolutely able of executing $t_k$) if $\Gamma_i \geq d(t_k)$. Analogously, $t_k$ is \emph{conditionally feasible} to $r_i$ (or $r_i$ is conditionally able of executing $t_k$) since $\gamma_i^j \geq d(t_k)$. In other words, robots are absolutely able of performing all picking tasks whose demand is less than or equal to their maximum load capacity; and conditionally able to perform any picking task whose demand is less than or equal to their current load capacity. By default, all robots are absolutely able of performing any delivery task, but not all robots are able of performing all picking tasks. Let $f_{fact}(\zeta_k, r_i)$ be a binary Boolean function that determines whether any task $\zeta \in T \cup H$ is feasible at $r_i$ after $r_i$ execute $\zeta_{k-1 }$, so mathematically:

\begin{align}
    f_{fact}(\zeta_k, r_i) = 
\begin{cases}
	\mbox{true},   & \mbox{if } \zeta_k \in H \\
	\mbox{true},   & \mbox{if } \zeta_k \in T \enskip \wedge \enskip \zeta_k \mbox{ is the first picking task } \enskip \wedge \enskip \Gamma_i \geq d(\zeta_k) \\
    \mbox{true},   & \mbox{if } \zeta_k \in T \enskip \wedge \enskip \gamma_i^{k-1} \geq d(\zeta_k) \\
    \mbox{false},  & \mbox{if } \zeta_k \in T \enskip \wedge \enskip (\Gamma_i < d(\zeta_k) \enskip \vee \enskip \gamma_i^{k-1} < d(\zeta_k))
\end{cases}
\end{align}

Let $\theta_i^{*}$ be the set of all absolutely feasible picking tasks to $r_i$, $\theta_i^{t_j}$ the set of all conditionally feasible picking tasks to $r_i$ after $r_i$ executes $t_j$ and $\tau_i = (\tau_{i:1}, \ldots, \tau_{i:q})$ an ordered sequence of $q$ picking tasks assigned to $r_i$. So, a feasible sequence is every sequence $\tau_k$ such that:

\begin{description}
	\item $\tau_{k:1} \in \theta_k^{*}$;
	\item $\tau_{k:j} \in \theta_k^{\tau_{k:j-1}}$, for $2 \leq j \leq q$;
	\item $\sum_{\tau_{k:j} \in \tau_k} d(\tau_{k:j}) \leq \Gamma_k$
\end{description}

In other words, $\tau_k$ is a feasible sequence if (i) the first task is absolutely feasible to $r_k$; if (ii) the other tasks are conditionally feasible to $r_k$ after $r_k$ executes the previous task and if (iii) the sum of all tasks' demand in $\tau_k$ does not exceed the $r_k$'s maximum capacity.

Such a definition of a feasible sequence implies that robots must restore the maximum load capacity between two sequences, which only occurs after a delivery task. As our goal is to reduce costs, we determine that every delivery task after a feasible sequence $\tau$ must be the one whose delivery station place is closest to the last task of $\tau$.

Now, let $A_i = (\alpha_{i:1}, \ldots, \alpha_{i:a})$ be a route containing picking and delivery tasks assigned to $r_i$. $A_i$ is only a feasible route if :

\begin{align}
	A_i = \tau_i \cup \{\beta\}
\end{align}

\noindent
where:
\begin{enumerate}
	\item $\tau_i$ is a feasible sequence that makes up $A_i$; and
	\item $\{\beta\}$ is a unitary set whose element $\beta$ is the delivery task that represents the closest delivery station to $\tau_{i:q}$ (the last task of $\tau_i$).
\end{enumerate}

Note that $A_i$ is one of the $t$ routes of $r_i$ that make up the ideal solution of HFMDTA-DR (Section \ref{subsec:baseline_formulation_heterogeneous_dispersed}). Each task $\alpha_{i:k} \in A_i$ comprises traversing the edge connecting the corresponding vertices in the graph $G^i$. The first task comprises traversing the edge that connects the vertex $\alpha_{i:1}$ (or $v_0^i$) with the vertex that corresponds to the first task. Thus, the cost $k_i^{v \rightarrow u}$ of traveling from the vertex $v$ to the vertex $u$ is equal to the weight of the edge $(v, u) \in E^i$, i.e.:

\begin{align}
	k_i^{v \rightarrow u} = cost(v, u, \upsilon_i)
\end{align}

\noindent
where $\upsilon_i$ is the traffic speed of $r_i$.

Let $\Delta_v^u$ be the Manhattan distance between $v$ and $u$, then the cost $k_i^{v \rightarrow u}$ of $r_i$ to travel from $v$ to $u$ is given by Equation \ref{math:costs}.

\begin{align}
    cost(v, u, \upsilon_i) =
\begin{split}
\begin{cases}
    \infty,   & \mbox{if } u \in T \enskip \wedge \enskip u \notin \theta_i^{*} \\
	\frac{\Delta_v^u}{\upsilon_i},   & \mbox{if } u \in H \enskip \vee \enskip (u \in T \enskip \wedge \\ 
	                                 & \enskip u \in \theta_i^{*} \enskip \wedge \enskip  f_{fact}(u, r_i) = \mbox{true} \enskip  \wedge \\
									 & \enskip \exists c \in T \enskip \mid \enskip c \neq u, f_{fact}(c, r_i)) = \mbox{true} \\
    \frac{\Delta_v^u}{\upsilon_i} + \frac{\Delta_u^{I_u}}{\upsilon_i},   & \mbox{if } u \in T \enskip \wedge \enskip u \in \theta_i^{*} \enskip \wedge \\
	                                 & \enskip f_{fact}(u, r_i) = \mbox{true} \enskip \wedge \\
									 & \enskip \forall c \in T \enskip \mid \enskip c \neq u, f_{fact}(c, r_i) = \mbox{false} \\
	\frac{\Delta_v^{I_v}}{\upsilon_i} + \frac{\Delta_{I_v}^u}{\upsilon_i},   & \mbox{if } u \in T \enskip \wedge \enskip u \in \theta_i^{*} \enskip \wedge \\
	                                 & \enskip f_{fact}(u, r_i) = \mbox{false} \enskip \wedge \\
									 & \enskip \exists c \in T \enskip \mid \enskip c \neq u, f_{fact}(c, r_i) = \mbox{true} \\
    \frac{\Delta_v^{I_v}}{\upsilon_i} + \frac{\Delta_{I_v}^u}{\upsilon_i} + \frac{\Delta_u^{I_u}}{\upsilon_i},   & \mbox{if } u \in T \wedge u \in \theta_i^{*} \enskip \wedge \\
	                                 & \enskip f_{fact}(u, r_i) = \mbox{false} \enskip \wedge \\
									 & \forall c \in T \enskip \mid \enskip c \neq u, f_{fact}(c, r_i) = \mbox{false} \\
\end{cases}
\end{split}
\label{math:costs}
\end{align}

Equation \ref{math:costs} depends on five different cases:

\begin{enumerate}
	\item \textbf{Case 1:} the vertex $u$ corresponds to a picking task ($u \in T$), but the $r_i$'s maximum load capacity is insufficient to execute $u (u \notin \theta_i^{ *})$, so $cost(v, u, \upsilon_i) = \infty$;
	\item \textbf{Case 2:}
	\begin{enumerate}
		\item the vertex $u$ corresponds to a delivery task ($u \in H$) or;
		\item the vertex $u$ corresponds to a picking task ($u \in T$), the $r_i$'s maximum and current load capacity is sufficient to execute $u (u \in \theta_i^{*}$ and $f_ {fact}(u, r_i) = \mbox{true})$ and $r_i$ do not need to visit a delivery station after executing $u$, as there will be at least one feasible task besides $u$ $(\ exists \enskip c \in T \enskip \mid \enskip c \neq u, f_{fact}(c, r_i) = \mbox{true})$;
	\end{enumerate}
	\item \textbf{Case 3:} the vertex $u$ corresponds to a picking task ($u \in T$), the $r_i$'s maximum and current capacity is sufficient to execute $u (u \in \theta_i^ {*}$ and $f_{fact}(u, r_i) = \mbox{true})$ and $r_i$ need to visit a delivery station after executing $u$, as all other tasks besides $u$ do not will be feasible to $r_i (\forall c \in T \enskip \mid \enskip c \neq u, \neg f_{fact}(c, r_i) = \mbox{false})$;
	\item \textbf{Case 4:} vertex $u$ corresponds to a picking task ($u \in T$), the $r_i$'s current capacity is insufficient to execute $u (f_{fact}(u, r_i) = \mbox {false})$, even though the maximum capacity is enough to perform $u (u \in \theta_i^{*})$. Also, $r_i$ does not need to visit a delivery station after executing $u$, as there will be at least one 	feasible task besides $u$ $(\exists c \in T \enskip \mid \enskip c \neq u, f_{fact}(c, r_i) = \mbox{true})$;
	\item \textbf{Case 5:} vertex $u$ corresponds to a picking task ($u \in T$), the $r_i$'s current capacity is insufficient to execute $u (f_{fact}(u, r_i) = \mbox {false})$, even though the maximum capacity is enough to perform $u (u \in \theta_i^{*})$. Also, $r_i$ needs to visit a delivery station after executing $u$, as all other tasks besides $u$ will not be feasible to $r_i (\forall c \in T \enskip \mid \enskip c \neq u , f_{fact}(c, r_i) = \mbox{false})$;
\end{enumerate}

Generally, cost estimators conceived in other works only take into account the cost between the current location and the target location, without considering the possibility that the robots will not have enough load capacity to perform the target task at any given time. Our cost estimator contrasts the more common models because it was designed to penalize visits to delivery stations and thus optimize the robots' usage and reduce costs.

DoNe-CPTA is a task-allocation heuristic that applies the proposed cost estimator to create action groups, optimizing the allocation of tasks to a team of heterogeneous robots, and reducing visits to delivery stations. Section \ref{sec:algorithm} presents the details of this novel heuristic to solve HFMDTA-DR in smart warehouses.
\section{Algorithm Design}
\label{sec:algorithm}

DoNe-CPTA is a task allocation algorithm guided by an adaptation of the Voronoi Tessellation \cite{Pokojski2018141}, whose definition is as follows:

\begin{definition}[Voronoi Tesselation]
Let $M$ be a set of points in a given 2D space and $X = \{x_1, \ldots, x_t\}$ a subset of $t$ points in $M$ called \emph{sites}. All points in $P$, where $P \enskip \cup \enskip X = M$, make up the \emph{tessellation space} in $M$. Voronoi Tesselation assigns each point in $P$ to its nearest site in $X$. The \emph{site cell} $x_i$ is the set of all points in $P$ that are closer to $x_i$ than to any site $x_j$, where $i$ is any index different than $j$.
\end{definition}

Concerning DoNe-CPTA, sites and tessellation space characterize robots and tasks, respectively. A robot cell corresponds to tasks that a given robot can perform at a lower cost than any other robot. Our cost estimator (Equation \ref{math:costs}) is used by DoNe-CPTA to compute all robot cells.

Formally, $\Phi[r_k] \subseteq T$ symbolizes the set of all tasks that $r_k$ can perform at the lowest cost among all robots in $R$. Similarly, $\Psi[t_k] \subseteq R$ symbolizes the set of all robots that execute $t_k$ at the lowest cost. $\Phi[r_k]$ we also call $r_k$'s \emph{domain} and $\Psi[t_k]$ we call $t_k$'s \emph{dominants}. Mathematically:

\begin{align}
	\label{algn:phi}
	\Phi[r_k] = \{t_j \in T \enskip \mid \enskip \kappa_k^{P[r_k] \rightarrow P[t_j]} < \kappa_i^{P[r_i] \rightarrow P[t_j]}, \enskip \forall \enskip r_i \neq r_k\}
\end{align}

\begin{align}
	\label{algn:psi}
	\Psi[t_k] = \{r_j \in R \enskip \mid \enskip \kappa_j^{P[r_j] \rightarrow P[t_k]} < \kappa_i^{P[r_i] \rightarrow P[t_k]}, \enskip \forall \enskip r_i \neq r_j\}
\end{align}

Following Definitions \ref{def:phi} and \ref{def:psi} are consequences of \ref{algn:phi} and \ref{algn:psi}, respectively:

\begin{definition}
	\label{def:phi}
	Robot $r_k$ dominates task $t_j$ if and only if $t_j \in \Phi[r_k]$;
\end{definition}

\begin{definition}
	\label{def:psi}
	Robot $r_j$ is a $t_k$'s dominant if and only if $r_j \in \Psi[t_k]$.
\end{definition}

Note that cases $4$ and $5$ of equation \ref{math:costs} computes costs for conditionally unfeasible tasks allowing such tasks to make up $\Phi[r_i]$ occasionally. This particular case means that if $r_i$ executes a delivery task before (or after) the respective picking task, then the estimated cost will be less than executing such a picking task directly.

In the first step, DoNe-CPTA computes the domain of all robots. Then, each robot receives the lowest cost task within its domain. Not all robots can receive picking tasks either for not dominating any task or because the robot's variable load capacity is less than the lowest cost task's demand. Then, DoNe-CPTA updates the robot's position to the position of its last task. If any robot $r_i$ receives a picking task, our algorithm decreases the task's demand from $r_i$'s variable load capacity. Otherwise, $r_i$ resets its variable load capacity to the maximum load capacity. $\Phi$ and $\Psi$ must be constantly updated because both are computed as a function of the robots' location and capacity. DoNe-CPTA continues as long as unassigned picking tasks exist.

Algorithm \ref{alg:domain_comput} shows the pseudo-code of the domain computation algorithm (\emph{computeDomain}). $\psi$ (line \ref{cptDomainAlgln2}) is a temporary variable that stores all robots references that dominate $t_j$ (line \ref{cptDomainAlgln1}). \emph{computeDomain} iterates over all robots in $R$ (line \ref{cptDomainAlgln4}) for each task in $T$. $\kappa_{current}$ stores the current robot-task pair's cost according to \ref{math:costs} (line \ref{cptDomainAlgln5}).

\begin{algorithm}
\caption{computeDomain Algorithm.}
\label{alg:domain_comput}
\hspace*{\algorithmicindent} \textbf{Input:} $R$, $T$\\
\hspace*{\algorithmicindent} \textbf{Output:} $\Phi$, $\Psi$
\begin{algorithmic}[1]
\ForEach {$t_j \in T$}\label{cptDomainAlgln1}
	\State $\psi \Leftarrow \emptyset$\label{cptDomainAlgln2}
	\State $\kappa_{best} \Leftarrow \infty$\label{cptDomainAlgln3}
	\ForEach {$r_i \in R$}\label{cptDomainAlgln4}
		\State $\kappa_{current} \Leftarrow \kappa_i^{P[t_j] \rightarrow P[r_i]}$\label{cptDomainAlgln5}
		\If{$\kappa_{current} = \kappa_{best}$}\label{cptDomainAlgln6}
			\State $\psi \Leftarrow \psi \cup {r_i}$\label{cptDomainAlgln7}
		\ElsIf {$\kappa_{current} < \kappa_{best}$}\label{cptDomainAlgln8}
			\State $\kappa_{best} \Leftarrow \kappa_{current}$\label{cptDomainAlgln9}
			\State $\psi \Leftarrow \emptyset$\label{cptDomainAlgln10}
			\State $\psi \Leftarrow \psi \cup r_i$\label{cptDomainAlgln11}
		\EndIf\label{cptDomainAlgln12}
	\EndFor\label{cptDomainAlgln13}
	\ForEach {$r_i \in \psi$}\label{cptDomainAlgln14}
		\State $\Phi[r_i] \Leftarrow t_j$\label{cptDomainAlgln15}
	\EndFor\label{cptDomainAlgln16}
	\State $\Psi[t_j] \Leftarrow \psi$\label{cptDomainAlgln17}
\EndFor\label{cptDomainAlgln18}
\end{algorithmic}
\end{algorithm}

From line \ref{cptDomainAlgln6} to \ref{cptDomainAlgln7}, \emph{computeDomain} includes $r_i$ in $\psi$ if $r_i$-$t_j$ pair's cost is equal to the cost of robots that are already in $\psi$. Multiple robots can dominate a single task with equal costs. Condition in line \ref{cptDomainAlgln8} will be true if the computed cost in line \ref{cptDomainAlgln5} is less than those already computed. In the latter case, \emph{computeDomain} resets $\psi$ and then append $r_i$ to it again (lines from \ref{cptDomainAlgln9} to \ref{cptDomainAlgln11}).

After computing $\Psi[t_j]$, the last loop (lines \ref{cptDomainAlgln14} to \ref{cptDomainAlgln16}) inserts $t_j$ as a task that is dominated by all robots in $\psi$. The domain algorithm outputs $\Phi$ and $\Psi$. Each $\Phi$ element is the set of tasks dominated by $r_i$ ($\Phi[r_i]$) and each $\Psi$ element is the set of robots that dominate $t_j$ ($\Psi[t_j ]$).

$\Psi$ and $\Phi$ are valid sets as long as robots' position and capacity remain the same as when both sets were computed. However, as DoNe-CPTA iteratively adjusts robots' position and capacity, $\Phi$ becomes invalid when robots receive new tasks. Likewise, all $\Psi$ sets originating from $\Phi$ also become invalid. In this scenario, our strategy requires that robot domains are constantly updated.

Therefore, we propose a simulation-based approach to determine domain recomputation periods and prevent DoNe-CPTA from assigning lower-cost tasks from an invalid domain. Such a simulation employs an arrival queue (AQ) that stores the time robots will take to reach their latter tasks. AQ contains a slot for each robot, and each slot stores information about the robot itself and the arrival time until its next task. Therefore, robots' movements are simulated by updating the arrival times in their respective slots. We use the information stored in the $r_i$ slot to test the validity of $\Phi[r_i]$: if $r_i$ is not executing tasks and its respective time is zero, then $\Phi [r_i]$ is valid. Otherwise, $\Phi[r_i]$ is invalid.

Initially, AQ holds zeros for all robots, as they are all idle. DoNe-CPTA always selects the first robot in AQ. If the robot's domain is valid, it will receive the lowest cost picking task within its domain (or a delivery task, depending on its load capacity). After that, DoNe-CPTA updates the corresponding slot in AQ. Finally, the queue is sorted in ascending order by arrival times.

However, if the selected robot's domain is invalid, DoNe-CPTA recalculates the domain of all robots. Such a robot is removed from the queue and only reinserted again when it receives a new task. Queue times are maintained, and the DoNe-CPTA selects the next robot after recalculating all domains. Algorithm \ref{alg:done_cpta} shows the pseudo-code of DoNe-CPTA, whose inputs are the sets $R$, $T$, and $H$. Lines from \ref{DoNeCPTAAlgln1} to \ref{DoNeCPTAAlgln3} initialize the output variables, where $\lambda$ represents the set of used robots.

\begin{algorithm}
\caption{DoNe-CPTA Algorithm.}
\label{alg:done_cpta}
\hspace*{\algorithmicindent} \textbf{Input:} $R$, $T$, $H$\\
\hspace*{\algorithmicindent} \textbf{Output:} Each robot $r_i \in R$ gets a route $A_i$, $cost$, $nDepots$, $usedRobots$
\begin{algorithmic}[1]
\State $cost \Leftarrow 0$\label{DoNeCPTAAlgln1}
\State $nDepots \Leftarrow 0$\label{DoNeCPTAAlgln2}
\State $\lambda \Leftarrow \emptyset$\label{DoNeCPTAAlgln3}
\While {$T \neq \emptyset$}\label{DoNeCPTAAlgln4}
	\State [$\Phi$, $\Psi$] $\Leftarrow$ computeDomain($R$, $T$)\label{DoNeCPTAAlgln5}
	\State prepareAQ($R$)\label{DoNeCPTAAlgln6}
	\While {true}\label{DoNeCPTAAlgln7}
		\If {!validDomain()}\label{DoNeCPTAAlgln8}
		    \State \textbf{break}\label{DoNeCPTAAlgln9}
		\EndIf\label{DoNeCPTAAlgln10}
		\State $r_{current} \Leftarrow$ nextRobot()\label{DoNeCPTAAlgln11}
		\State decreaseArrivalTime($r_{current}$, $R$)\label{DoNeCPTAAlgln12}
		\State $\varphi_{r_{current}} \Leftarrow \Phi_(r_{current})$\label{DoNeCPTAAlgln13}
		\If {$\varphi_{r_{current}} \neq \emptyset$}\label{DoNeCPTAAlgln14}
			\State $t_{current} \Leftarrow \arg\min_{t \in \varphi_{r_{current}}} \kappa_i^{P[r_i] \rightarrow P[t]}$\label{DoNeCPTAAlgln15}
			\If {$r_{current}.currentCapacity < t_{current}.demand$}\label{DoNeCPTAAlgln16}
				\State $h_{best} \Leftarrow$ bestDepot($H$)\label{DoNeCPTAAlgln17}
				\State $A_{current} \Leftarrow A_{current} \cup h_{best}$\label{DoNeCPTAAlgln18}
				\State $cost \Leftarrow cost + \kappa_i^{r_{current} \rightarrow h_{best}}$\label{DoNeCPTAAlgln19}
				\State $nDepots \Leftarrow nDepots + 1$\label{DoNeCPTAAlgln20}
				\State resetCurrentCapacity($r_{current}$)\label{DoNeCPTAAlgln21}
				\State setArrivalTime($r_{current}$, $h_{best}$)\label{DoNeCPTAAlgln22}
			\Else\label{DoNeCPTAAlgln23}
				\State $A_{current} \Leftarrow A_{current} \cup t_{current}$\label{DoNeCPTAAlgln24}
				\State $cost \Leftarrow cost + \kappa_i^{r_{current} \rightarrow t_{current}}$\label{DoNeCPTAAlgln25}
				\State setCurrentCapacity($r_{current}$)\label{DoNeCPTAAlgln26}
				\State $T \Leftarrow T \backslash \{t_{current}\}$\label{DoNeCPTAAlgln27}
				\State $\lambda \Leftarrow \lambda \cup {r_{current}}$\label{DoNeCPTAAlgln28}
				\State setArrivalTime($r_{current}$, $t_{current}$)\label{DoNeCPTAAlgln29}
				\State setAsNotValid($\Phi - \Phi[r_{current}]$, $\Psi[t_{current}]$)\label{DoNeCPTAAlgln30}
			\EndIf\label{DoNeCPTAAlgln31}
			\State setCurrentPosition($r_{current}$)\label{DoNeCPTAAlgln32}
		\EndIf\label{DoNeCPTAAlgln33}
	\EndWhile\label{DoNeCPTAAlgln34}
\EndWhile\label{DoNeCPTAAlgln35}
\State $[lastDepotsCost, lastDepotsN] \Leftarrow$ addLastDepots($R$, $nDepots$)\label{DoNeCPTAAlgln36}
\State $cost \Leftarrow cost + lastDepotsCost$\label{DoNeCPTAAlgln37}
\State $nDepots \Leftarrow nDepots + lastDepotsN$\label{DoNeCPTAAlgln38}
\State $usedRobots \Leftarrow \vert\lambda\vert$\label{DoNeCPTAAlgln39}
\end{algorithmic}
\end{algorithm}

DoNe-CPTA stop condition is found in line \ref{DoNeCPTAAlgln4}. The main iteration takes place while unassigned picking tasks exist. The first step is to calculate the domain of all robots based on their distribution and tasks across the map and their different capacities and demands. This is done by the function \texttt{domainOfR($R$, $T$)}, according to \emph{computeDomain} (line \ref{DoNeCPTAAlgln5}).

\verb+prepareAQ(R)+ (line \ref{DoNeCPTAAlgln6}) initializes with zero all slots in AQ. If AQ has already been initialized in another iteration, then \texttt{prepareAQ($R$)} conserves existing times and orders robots by arrival time. If two or more robots have the same arrival time, they are sorted in descending order by the number of tasks they dominate.

The AQ is updated all the time between lines \ref{DoNeCPTAAlgln8} and \ref{DoNeCPTAAlgln33}. The slot order dynamically changes depending on the distance and speed of each robot. \texttt{validDomain()} tests whether the domain of the first robot in AQ is valid. The nested loop only ends at line \ref{DoNeCPTAAlgln9} when \texttt{validDomain()} returns false. Then \texttt{nextRobot()} returns the first robot in AQ (line $11$).

\texttt{decreaseArrivalTime()} and \texttt{setArrivalTime()} constantly manipulate the AQ inside the nested loop. In \texttt{decreaseArrivalTime($r_{current}$, $R$)} (line \ref{DoNeCPTAAlgln12}) DoNe-CPTA takes the stored time of $r_{current}$ and uses such time to decrement all other robots' times in the queue. All robots whose time reaches zero will have their domains marked as invalid. The domain of $r_{current}$ is also marked invalid within this function.

\texttt{setArrivalTime()} (lines \ref{DoNeCPTAAlgln22} and \ref{DoNeCPTAAlgln30}) computes the time that the robot will take to reach the next task, sets this time in the $r_{current}$ slot, and sorts the arrival queue. On line \ref{DoNeCPTAAlgln13}, $\varphi$ receives all tasks dominated by $r_i$. If $r_i$ dominates some task (line \ref{DoNeCPTAAlgln14}) then $t_{current}$ gets the lowest cost task out of all the tasks in $\varphi$ (line \ref{DoNeCPTAAlgln15}).

The test in line \ref{DoNeCPTAAlgln16} checks if $t_{current}$ is not conditionally feasible to $r_{current}$. This is done by comparing the task demand and the robot's current capacity. If the test is true, it means that the robot must visit the nearest delivery station. \texttt{bestDepot($H$)} (line \ref{DoNeCPTAAlgln17}) returns the delivery task that represents the nearest station. The task is then appended into the $r_{current}$ route (line \ref{DoNeCPTAAlgln18}) and the cost and amount of visits to delivery stations are updated (lines \ref{DoNeCPTAAlgln19} and \ref{DoNeCPTAAlgln20}). The capacity is then reset to the robot's maximum capacity (line \ref{DoNeCPTAAlgln21}). At line \ref{DoNeCPTAAlgln22}, \texttt{setArrivalTime($r_{current}$, $h_{best}$)} is called to update the $r_{current}$ slot in AQ.

If the test on line \ref{DoNeCPTAAlgln16} is false, then the robot will be able to run $t_{current}$. At line \ref{DoNeCPTAAlgln24}, $t_{current}$ is appended into the $r_{current}$ route. The cost and current capacity of the robot are also updated (rows \ref{DoNeCPTAAlgln25} and \ref{DoNeCPTAAlgln26}). Also, $t_{current}$ is removed from $T$ and $r_{current}$ is appened in $\lambda$. \texttt{setArrivalTime($r_{current}$, $t_{current}$)} updates $r_{current}$ slot in AQ.

If $t_{current}$ is dominated by other robots, then removing it from $T$ makes the domain of such robots invalid. Then, \texttt{setAsNotValid($\Phi - \Phi[r_{current}]$, $\Psi[t_{current}]$)} (line \ref{DoNeCPTAAlgln30}) cycles through $\Psi[t_{current}]$ to find all robots that dominate over $t_{current}$ and mark the corresponding $\Phi$ as invalid. On line \ref{DoNeCPTAAlgln32}, DoNe-CPTA updates the $r_{current}$'s position to  the newly assigned task's position.

Outside the main loop, \texttt{addLastDepots($R$, $nDepots$)} (line \ref{DoNeCPTAAlgln36}) appends delivery tasks to each generated route, to ensure that all picking tasks are delivered. This function returns values that increment the total cost (line \ref{DoNeCPTAAlgln37}) and the number of delivery station visits (line \ref{DoNeCPTAAlgln38}). In line \ref{DoNeCPTAAlgln39}, $usedRobots$ stores the number of used robots. DoNe-CPTA outputs routes of each robot, the total cost of all routes, the number of visits to delivery stations, and the number of used robots.
\section{Datasets}
\label{sec:datasets}

Research on MRTA in smart warehouses has grown every year, and new solutions are constantly being proposed. However, to the best of our efforts we could not find a benchmark considering all the aspects we aim to study with the proposed work, all very plausible to smart warehouse scenarios. Overall, environments and task distributions are randomized, and considering the fact that the code of many algorithms is almost always closed and not available to the public, it becomes difficult to replicate and reproduce the results for an adequate comparison.

As a first step to solving this issue, we have developed and made available a new dataset for testing developed solutions for MRTA in smart warehouses. Authors can use this novel dataset to validate their work on operations research and other MRTA scenarios. The instances we have created are based on datasets widely used in operations research works \cite{CVRP2021}. We use instances of a dataset designed to cover several characteristics of real applications, called dataset X \cite{Uchoa20161}. Also, dataset X is ideal for our adaptations because it is a CVRP, which sets up a very simple reduction of HFMDVRP-DV (Section \ref{sec:problem_formulation}).

From now on, we call dataset X as \emph{base-dataset}. Each instance of base-dataset defines tasks and robots' location, tasks' demand, and a common load capacity for all robots. So, inserting information that characterizes the HFMDVRP-DV, we produced other four datasets: XMT, WMT, SMT, and RMT, which present variations regarding the robot fleet (homogeneous or heterogeneous), number of delivery tasks, and robots' dispersion. More specifically, RMT, SMT, WMT, and XMT:

\begin{enumerate}
	\item Employ heterogeneous robots based on real robots' specifications;
	\item Contains additional delivery tasks;
	\item Randomize the robots' starting position;
	\item Allow costing in the function of the Manhattan distance and the robots' traffic speed.
\end{enumerate}

Table \ref{tab:datasets_comparison} shows the comparison between the novel datasets. Each dataset instance is named as $[RMT \mid SMT \mid WMT \mid XMT]-t<m>-r<n>-d<p>$, where $<m>$ is the number of tasks, $<n>$ is the number of robots, and $<p>$ is the number of delivery stations.

For example, instances \texttt{RMT-t181-r23-d1}, \texttt{SMT-t181-r23-d4}, \texttt{WMT-t181-r23-d4} and \texttt{XMT-t181-r23- d1} are adaptations of the base-dataset \texttt{X-n181-k23}, whose number of picking tasks, robots and delivery tasks are $181$, $23$, and $1$ (or $4$), respectively.

\begin{table*}
\begin{center}
\begin{minipage}{\textwidth}
\caption{Comparison between our adapted datasets. XMT contains instances of the classic CVRP problem. RMT sets up an HFVRP problem because it contains CVRP properties (robots start and end in the same position), but with heterogeneous fleet. In WMT, homogeneous robots start and end at different positions (MDVRP-DV). SMT contains all HFMDVRP-DV properties.}
\label{tab:datasets_comparison}
\begin{tabular*}{\textwidth}{@{\extracolsep{\fill}}ccccc@{\extracolsep{\fill}}}
\toprule
Name & \begin{tabular}[c]{@{}c@{}}Problem\\ Type\end{tabular} & \begin{tabular}[c]{@{}c@{}}Robot\\ Fleet\end{tabular} & \begin{tabular}[c]{@{}c@{}}Robots' Starting\\ Position\end{tabular} & \begin{tabular}[c]{@{}c@{}}\# Delivery\\ Tasks\end{tabular} \\
\midrule
RMT           & HFVRP                                                            & Heterogeneous                                                  & \begin{tabular}[c]{@{}c@{}}Same of (single)\\ delivery task\end{tabular}       & 1                                                                    \\
SMT           & HFMDVRP-DV                                                      & Heterogeneous                                                  & \begin{tabular}[c]{@{}c@{}}Dispersed\\ on the map\end{tabular}                 & Many                                                                 \\
WMT           & MDVRP-DV                                                        & Homogeneous                                                    & \begin{tabular}[c]{@{}c@{}}Dispersed\\ on the map\end{tabular}                 & Many                                                                 \\
XMT           & CVRP                                                            & Homogeneous                                                    & \begin{tabular}[c]{@{}c@{}}Same of (single)\\ delivery task\end{tabular}       & 1                                                                    \\
\botrule
\end{tabular*}
\end{minipage}
\end{center}
\end{table*}

RMT and XMT contain properties of CVRP and HFVRP problems, respectively. Both datasets are formed by a single delivery station where robots must start and end their routes. XMT and RMT differ only by the fleet type (homogeneous or heterogeneous). WMT contains properties of MDVRP-DV problem (many delivery stations and dispersed homogeneous robots), while SMT employs features of  HFMDVRP-DV (many delivery stations and dispersed heterogeneous robots). All instances are costed depending on the Manhattan distance and robots' traffic speed.

All WMT and XMT instances contain generic robots, with load capacities and traffic speeds equal to each other. In each instance, robots have the same capacity as in the equivalent instance of the base dataset. RMT and SMT instances contain data about load capacity and traffic speed from several robots designed to transport items in many settings, including smart warehouses. Table 1 of the supplementary material \footnote{\url{https://github.com/geosoliveira/DoNe-CPTA/blob/main/Supplementary-material/Supplementary_material.pdf}} presents the information about all robots' load capacities and traffic speeds that make up the RMT and SMT datasets. The heterogeneous fleet that makes up each instance of such datasets is randomly defined based on the supplementary material's load capacity information presented in Table 1. Algorithm 1 of the supplementary material presents the pseudo-code of the algorithm that randomizes the heterogeneous fleets in both datasets.

In addition to robots, we have also inserted new delivery tasks into HFMDVRP-DV instances. The total of delivery tasks $p$ is proportional to the number of picking tasks, whose value is computed  by \ref{math:n_deliveryt}. The number of delivery tasks in XMT and RMT instances is one.

\begin{align}
	p = \ln m - 1
	\label{math:n_deliveryt}
\end{align}

All picking tasks' demands in new instances with homogeneous robots remained the same as in the base-dataset. However, we adapted the demands for datasets with heterogeneous robots to ensure that at least one robot had enough capacity to perform all tasks. The adapted demand $d_e^{'}(t_j)$ of $t_j$ in the instance $e$ of the \emph{datasets} for heterogeneous robots is given by the Equation \ref{math:adp_demands}

\begin{align}
	d_e^{'}(t_j) = \frac{\mu_e(\Gamma) \times d_e(t_j)}{C_e}
	\label{math:adp_demands}
\end{align}

\noindent
where $\mu_e(\Gamma)$ is the average of the robots' load capacities of the instance $e$ of the new dataset, $d_e(t_j)$ is the demand of the task $t_j$ in instance $e$ of the base dataset and $ C_e$ is the capacity of robots on instance $e$ of the base dataset.

The picking and delivery tasks' locations are the same for all datasets. In SMT and WMT, the position of all new delivery tasks was defined randomly, as many were generated. SMT and WMT also define all robots' positions randomly. All robots and tasks occupy a different point on the map, except in RMT and XMT instances, where robots start and end at the same location.
\myexternaldocument{6-datasets}

\section{Evaluation}
\label{sec:evaluation}

We evaluated the performance of DoNe-CPTA in three aspects: solution cost, number of robots, and algorithm execution time.
The solution cost is the sum of the costs of the routes assigned to the robots, according to Equation \ref{math:costs}. The cost of each route is the sum of the costs of all tasks on the route. As previously pointed out by the cost formulation, robots' starting and ending point, the number of visits to delivery stations, and the robots' speed influence the tasks' costs. Also the routes can include more than one visit to the delivery stations and a mandatory visit after the last picking task. The third performance metric is the task allocation algorithm execution time, in seconds, on a machine with the following configuration:

\begin{description}
	\item Intel Core i3-8130U processor with 4MB cache;
	\item 4GB DDR4 Memory;
	\item Windows 10 Pro 20H2;
\end{description}

We compared the performance of DoNe-CPTA with a state-of-the-art algorithm, adaptated to the proposed HFMDVRP-DV's scenario, which is represented by the datasets detailed on (Section \ref{sec:datasets}). According to its authors, the state-of-the-art \emph{nearest-neighbor based Clustering And Routing} (nCAR) algorithm (\cite{Sarkar20185022}) was designed to meet the requirements of smart warehouses and has been formulated based on one of the VRP variations, more specifically, the CVRP. We adapted nCAR to meet the HFMDVRP-DV's properties and named the novel algorithm as \emph{Adapted nCAR} (a-nCAR). Below, we briefly present the differences between a-nCAR and nCAR:

\begin{itemize}
	\item In nCAR, all robots depart from the same location, the location of the (single) delivery station. In a-nCAR, robots depart from different locations according to the novel dataset specifications.
	\item In each main iteration of nCAR, the algorithm creates a feasible route for each robot, as they all have the same capability. The lowest cost route is chosen. In a-nCAR, each robot's main iteration is performed, respecting its capabilities, and the chosen route is the one with the lowest cost.
	\item In nCAR, when robots receive a delivery task, it corresponds to visiting the (single) delivery station. In a-nCAR, when a robot receives a delivery task, it corresponds to visiting the closest delivery station to the last picking task of such a robot.
\end{itemize}

Note that such differences are designed to make nCAR suitable for serving different instances of CVRP. In other words, if the input instances meet the CVRP properties, then a-nCAR will work similarly to nCAR. We compared DoNe-CPTA with a-nCAR running dataset X, original CVRP formulation, and observed compatible cost routes generated in a shorter runtime. As our target scenario contains HFMDVRP-DV properties, we have omitted such results, making them available in the supplementary material. For the sake of reproducibility, the C/C++ code implementation of all algorithms and the experiments performed are available at \url{https://github.com/geosoliveira/DoNe-CPTA}.

First, Section \ref{subsec:evaluation1} presents DoNe-CPTA performances when performing WMT instances (multiple stations, homogeneous and dispersed robots) and SMT instances (multiple stations, heterogeneous and dispersed robots). Then, Section \ref{subsec:evaluation2} presents the performance of DoNe-CPTA when running XMT (one delivery station, homogeneous and non-dispersed robots) and RMT (one delivery station, heterogeneous and non-dispersed robots).

\subsection{Performance of DoNe-CPTA with Robots Dispersed on the Map}
\label{subsec:evaluation1}

In this section, we will evaluate the performance of Done-CPTA compared to a-nCAR running all instances of SMT and WMT. The first dataset (HFMDVRP-DV) includes the characteristics of the target scenario (multi-station delivery, heterogeneous, and dispersed robots), while the second dataset (MDVRP-DV) includes the same characteristics except for the heterogeneous fleet. Each dataset contains $100$ instances, and each instance contains a specific number of picking tasks, delivery tasks (i.e., delivery stations), and robots. All instances define stations and robots' positions, in addition to all robots that will compose the heterogeneous fleet (if SMT), according to dataset specifications in Section \ref{sec:datasets}. We created $30$ variations for each instance and ran each variation once. So, we had $30$ runs per instance, $3000$ runs per dataset, and $6000$ runs for configurations with dispersed robots. Since each instance is different, route costs, the execution time, and the number of robots varied in each execution. All instance variations of all datasets, including RMT and XMT, are available for download at \url{https://github.com/geosoliveira/VRP-Instances}

\subsubsection{Route Costs}
\label{subsubsec:evaluation1_cost_routes}

\begin{figure*}[ht]
	\centering
	\includegraphics[width=1\textwidth]{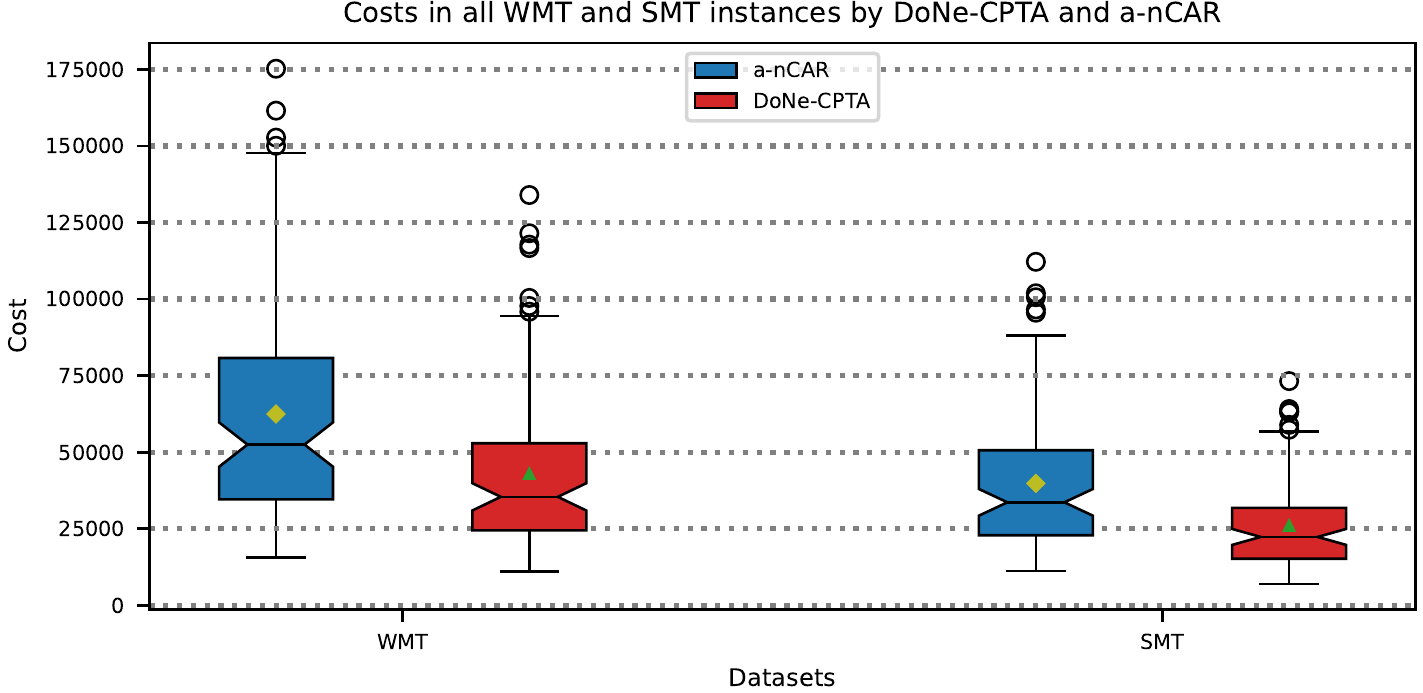}
	\caption{Overview of the cost of routes computed by DoNe-CPTA and a-nCAR when running SMT and WMT instances.}
	\label{fig:cost_dispersedrobots_boxplot}
\end{figure*}

Figure \ref{fig:cost_dispersedrobots_boxplot} shows an overview of routes' cost by DoNe-CPTA and a-nCAR. First of all, it is possible to see that the use of heterogeneous fleets favors the generation of lower-cost routes compared to the use of homogeneous fleets, regardless of the allocation algorithm. On average, DoNe-CPTA is more efficient than a-nCAR at a rate of $1.45$ using homogeneous fleets (Mann–Whitney U, $p = 0.00002624$), and $1.52$ using heterogeneous fleets (Mann–Whitney U, $p = 0.000002933$). Also, it is noteworthy that the performance of DoNe-CPTA using homogeneous robots is similar to the one of a-nCAR with heterogeneous robot (Mann-Whitney U, $p = 0.3797$). This interesting result demonstrates that the superior performance of the proposed algorithm can even compensate the unfavor scenario of using homogeneous fleet.


\begin{figure*}[ht]
	\centering
	\includegraphics[width=1\textwidth]{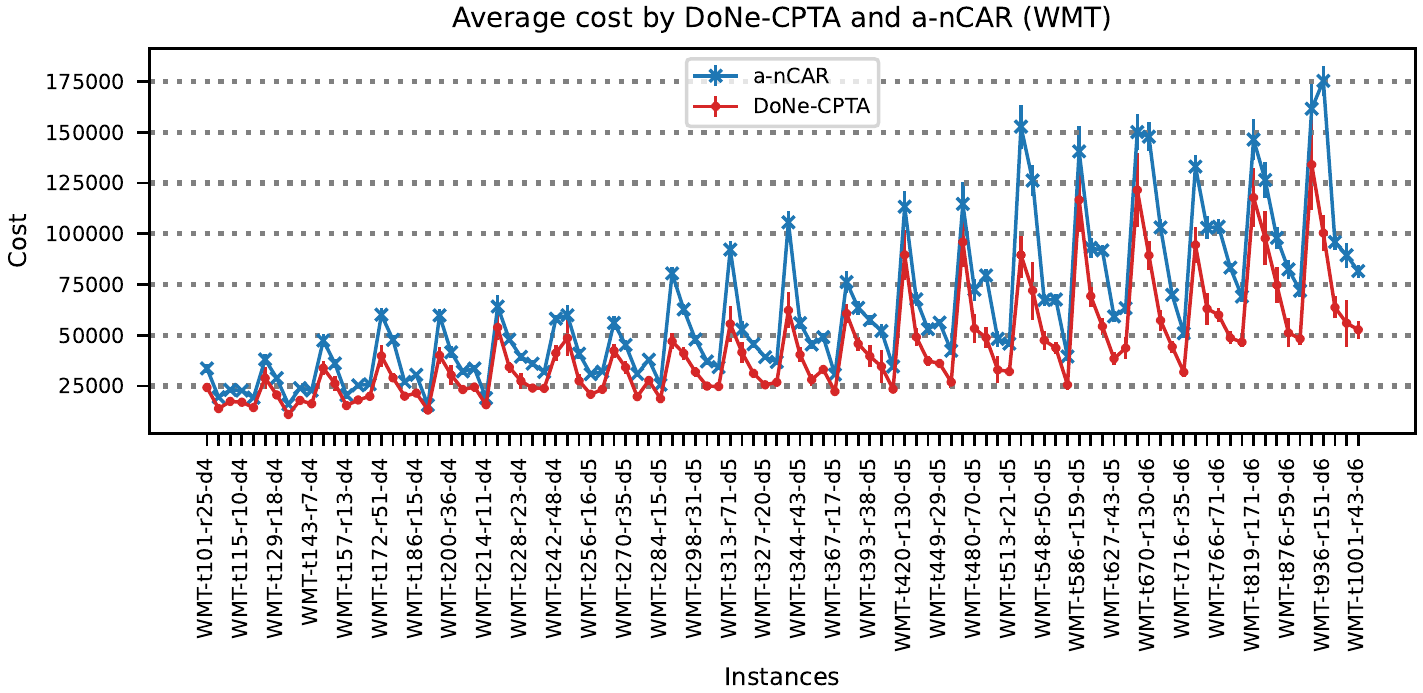}
	\caption{Cost of routes computed by DoNe-CPTA and a-nCAR with WMT.}
	\label{fig:dataset_wmt_cost}
\end{figure*}

\begin{figure*}[ht]
	\centering
	\includegraphics[width=1\textwidth]{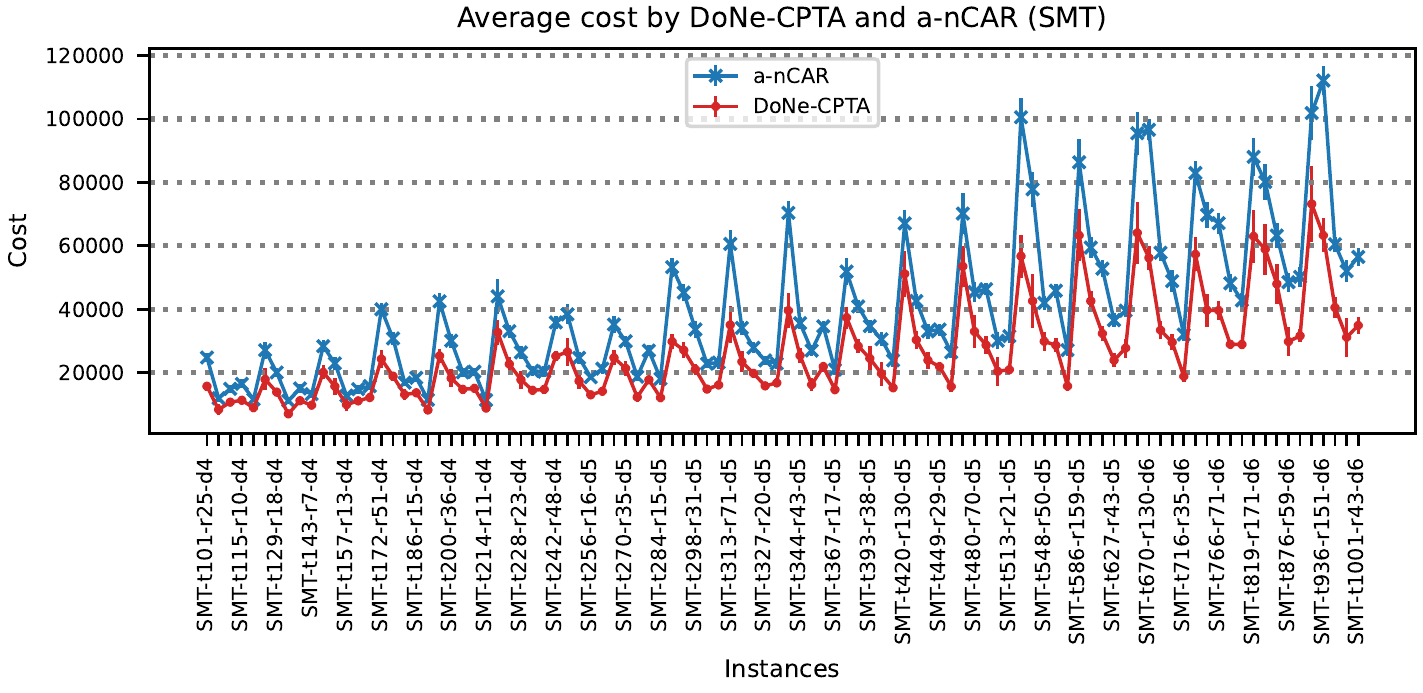}
	\caption{Cost of routes computed by DoNe-CPTA and a-nCAR with SMT.}
	\label{fig:dataset_smt_cost}
\end{figure*}

Figures \ref{fig:dataset_wmt_cost} and \ref{fig:dataset_smt_cost} shows the average cost of DoNe-CPTA and a-nCAR from the $30$ variations of WMT and SMT instances, respectively (the name of some instances were omitted for reasons of space, see the full result in the supplementary material\footnote{\url{https://github.com/geosoliveira/DoNe-CPTA/blob/main/Supplementary-material/Supplementary_material.pdf}}). As shown in Figure \ref{fig:cost_dispersedrobots_boxplot} DoNe-CPTA managed to generate lower-cost routes in all instances. Such costs are $33$\% and $30$\% lower on average in heterogeneous (SMT) and homogeneous(WMT) fleets, respectively. It is possible to see that the greater the number of the tasks the greater the performance advantage of DoNe-CPTA over a-nCAR, starting from $280$ tasks on. As expected, regardless of the dataset and strategy used, costs tend to increase when more robots are used.

Also, note that the costs of using a homogeneous and heterogeneous fleet are very different. Even with robots dispersed in equal locations, both algorithms achieve higher costs when dealing with robots with the same capacity load and traffic speed. This overall confirms the hypothesis that using a heterogenous and dispersed robots' fleet is advantageous in a smart warehouse scenario. Heterogeneous fleets cause allocation algorithms to compute lower-cost routes because robots with high traffic speed tend to have lower carrying capacity and vice versa. In this way, the fleet can supply its underlying disadvantages by appropriating each robot's advantages if the allocation algorithm considers the tasks' position and demand. On the other hand, homogeneous fleets with high-capacity robots will take longer to perform tasks, while those with high-speed robots will visit delivery stations more often, both making routes more costly.

In SMT (heterogeneous fleet), the highest cost routes occurred on instances with more than $900$ tasks, but similar costs occurred on instances with roughly $500$ in WMT (homogeneous fleet).

\subsubsection{Execution Time}
\label{subsubsec:evaluation1_exectime}

\begin{figure*}[ht]
	\centering
	\includegraphics[width=1\textwidth]{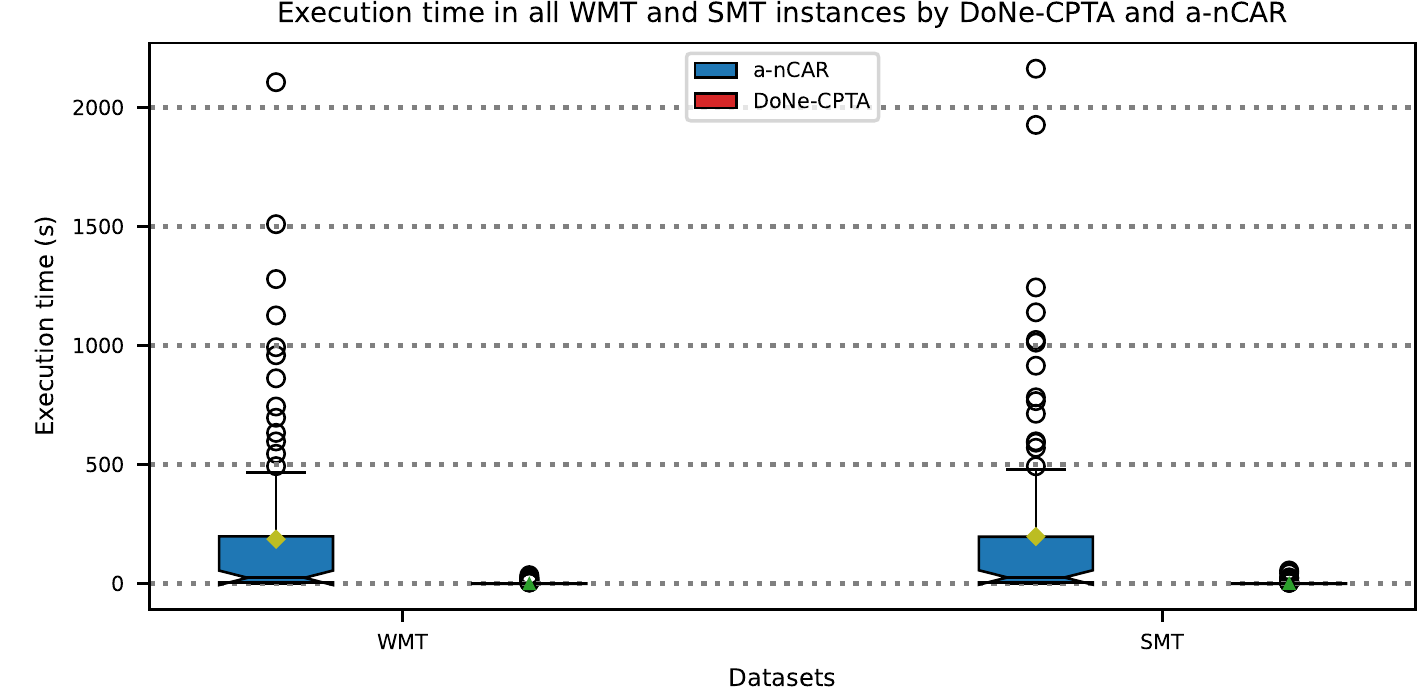}
	\caption{Overview of the execution time of DoNe-CPTA and a-nCAR when running SMT and WMT.}
	\label{fig:exectime_dispersedrobots_boxplot}
\end{figure*}

DoNe-CPTA is superior to a-nCAR also in terms of execution time, as can be seen in Figure \ref{fig:exectime_dispersedrobots_boxplot}. Most of the time, a-nCAR took from just under $1$ second to $200$ seconds to compute lower-cost routes. On some occasions, the execution time was longer than $1000$ seconds. This execution time might be prohibitive in highly dynamic environments where tasks are generated all the time, like a real-world smart warehouse. The average runtime of DoNe-CPTA is $3$ seconds with WMT and $4$ seconds with SMT. Additional and detailed information about the executions time can be found in the supplementary material\footnote{\url{https://github.com/geosoliveira/DoNe-CPTA/blob/main/Supplementary-material/Supplementary_material.pdf}}.


\begin{figure*}[ht]
	\centering
	\includegraphics[width=1\textwidth]{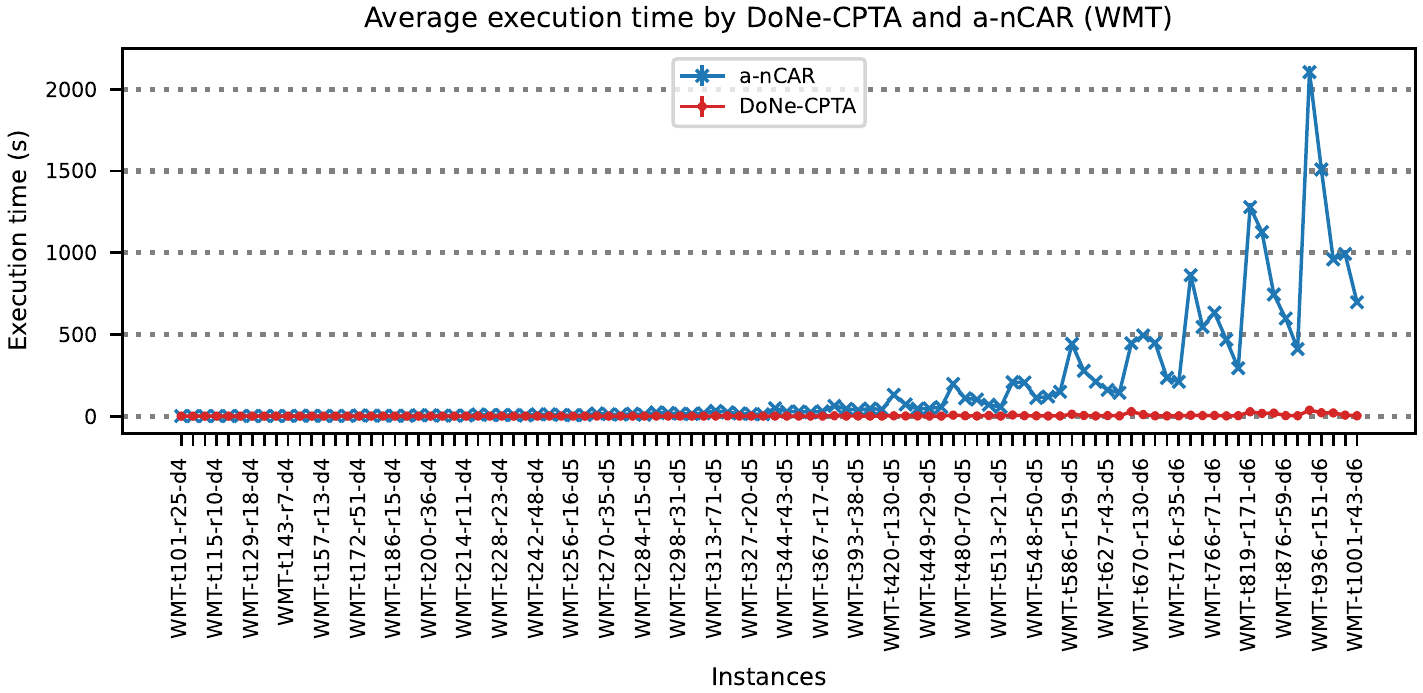}
	\caption{Execution times of a-nCAR and DoNe CPTA when performing WMT. Results represent an average of $30$ runs per instance, where each instance has different fleets of robots.}
	\label{fig:dataset_wmt_exec_time}
\end{figure*}

\begin{figure*}[ht]
	\centering
	\includegraphics[width=1\textwidth]{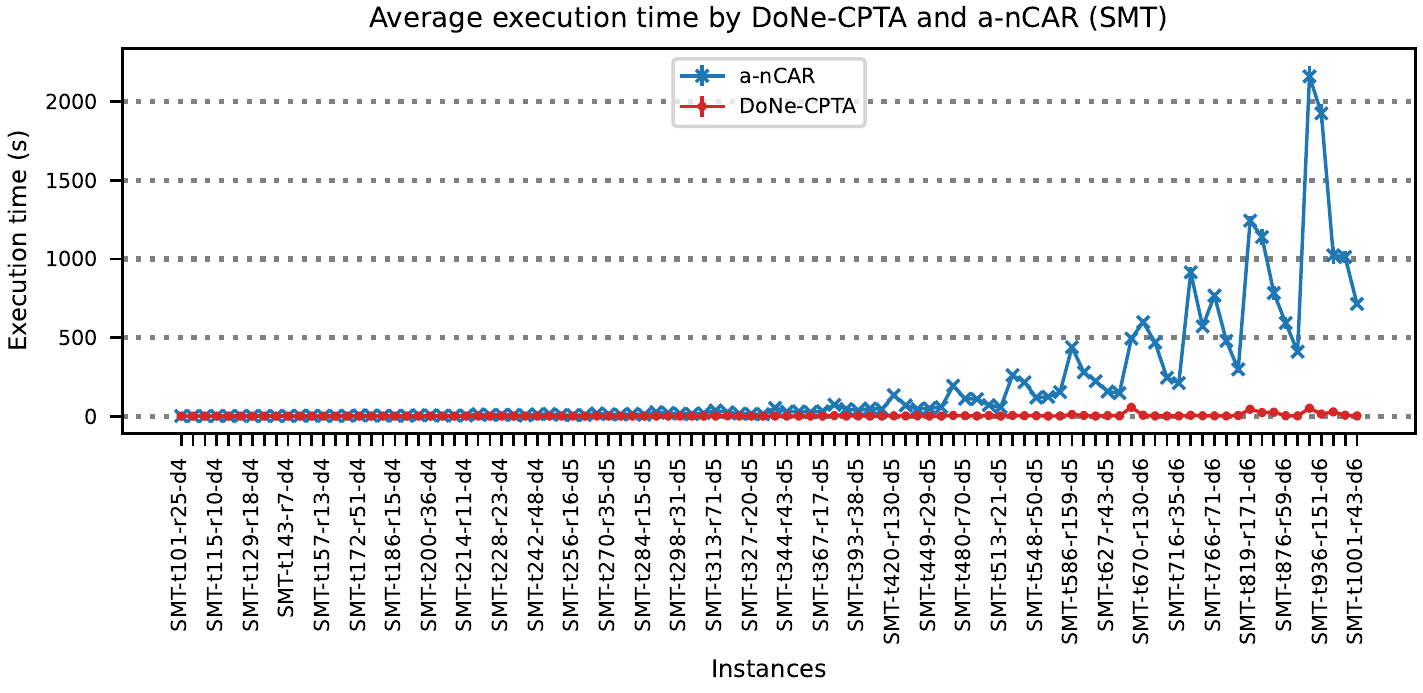}
	\caption{Execution times of a-nCAR and DoNe CPTA when performing SMT. Results represent an average of $30$ runs per instance, where each instance has different fleets of robots.}
	\label{fig:dataset_smt_exec_time}
\end{figure*}

Figures \ref{fig:dataset_wmt_exec_time} and \ref{fig:dataset_smt_exec_time} show the performance of DoNe-CPTA over a-nCAR in detail with the homogeneous and heterogeneous fleet datasets, respectively. Again, DoNe-CPTA was superior to a-nCAR in all instances regarding execution time. While a-nCAR took over $1000$ seconds to find solutions in some instances, our solution took $36$ and $51$ seconds in the most time-consuming instance of WMT and SMT, respectively. The difference becomes bigger as the number of tasks increases, and very noticeable from $420$ tasks on. As we show in Tables $2$ and $3$ of the supplementary material,
DoNe-CPTA was $96$\% more efficient in both datasets.

\subsubsection{Number of Robots}
\label{subsubsec:evaluation1_nrobots}

\begin{figure*}[ht]
	\centering
	\includegraphics[width=1\textwidth]{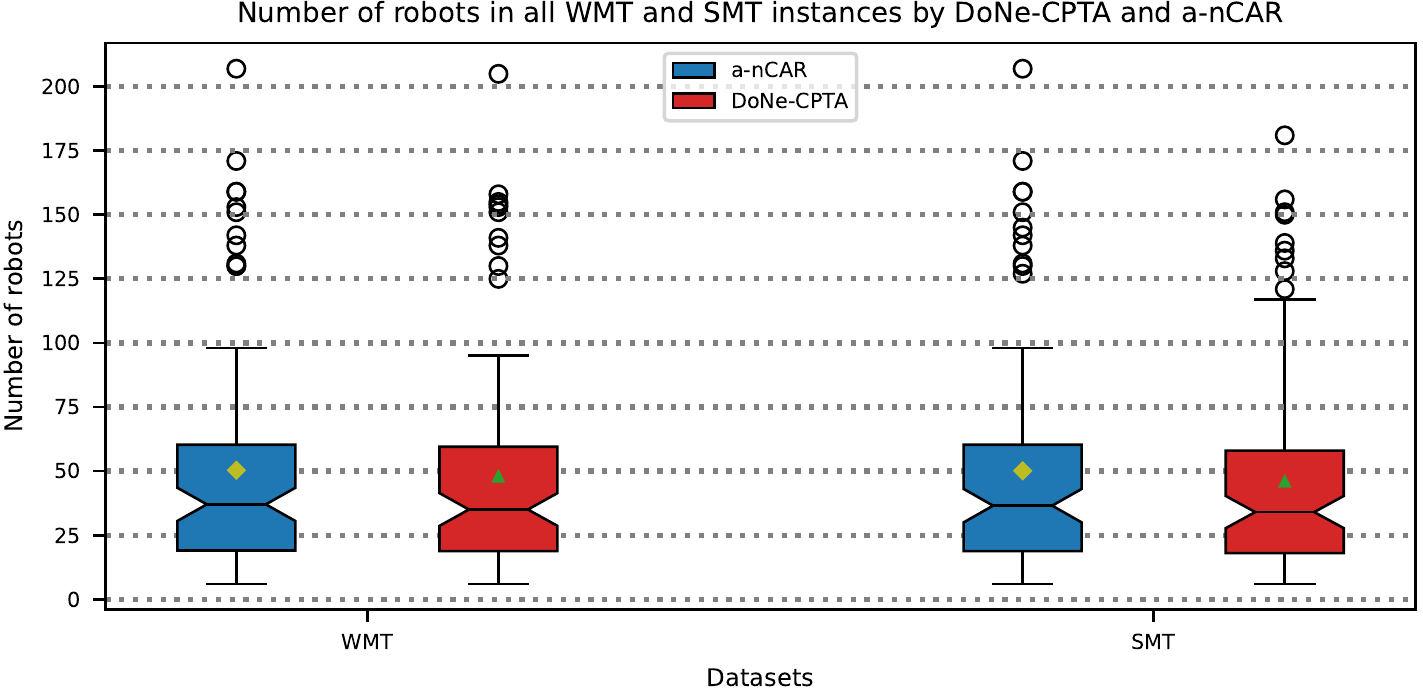}
	\caption{Overview of the number of robots by DoNe-CPTA and a-nCAR when running SMT and WMT.}
	\label{fig:nrobots_dispersedrobots_boxplot}
\end{figure*}

Figure \ref{fig:nrobots_dispersedrobots_boxplot} shows an overview of the number of robots employed by DoNe-CPTA and a-nCAR. We note that the type fleet influences route costs, but not the number of used robots. Virtually both Done-CPTA and a-nCAR employed $20$ to $60$ robots most of the time when executing SMT and WMT instances. In general, the number of robots tends to be the same regardless of the fleet and the used algorithm. However, DoNe-CPTA has shown itself to be more efficient in terms of the maximum number of robots. In other words,  employing AQ to control the arrival time of used robots saves robots not yet used without losing route cost efficiency.

Since DoNe-CPTA and a-nCAR tend to have similar performance regarding the number of robots used, we do not present detailed results for this metric in this section. However, all the results are included in the supplementary material.

\subsection{Performance of DoNe-CPTA with Robots in Fixed Location}
\label{subsec:evaluation2}

This section presents the performance of Done-CPTA with HFVRP problem instances (RMT) and with CVRP problem instances (XMT). Both datasets contain instances representing a scenario with a single station and robots starting at a fixed location, differing only in the fleet type. The RMT experimental setup was the same as the previous one: $100$ instances and $30$ variations, totaling $3000$ runs. There are no variations on XMT instances because the CVRP properties do not allow for fleet variation, multiple depots, and dispersed robots. Even so, each XMT instance ran $30$ times to compute the average execution of the algorithms. Therefore, the results we discussed in this section were taken from $6000$ runs.

\subsubsection{Route Costs}
\label{subsubsec:evaluation2_cost_routes}

\begin{figure*}[ht]
	\centering
	\includegraphics[width=1\textwidth]{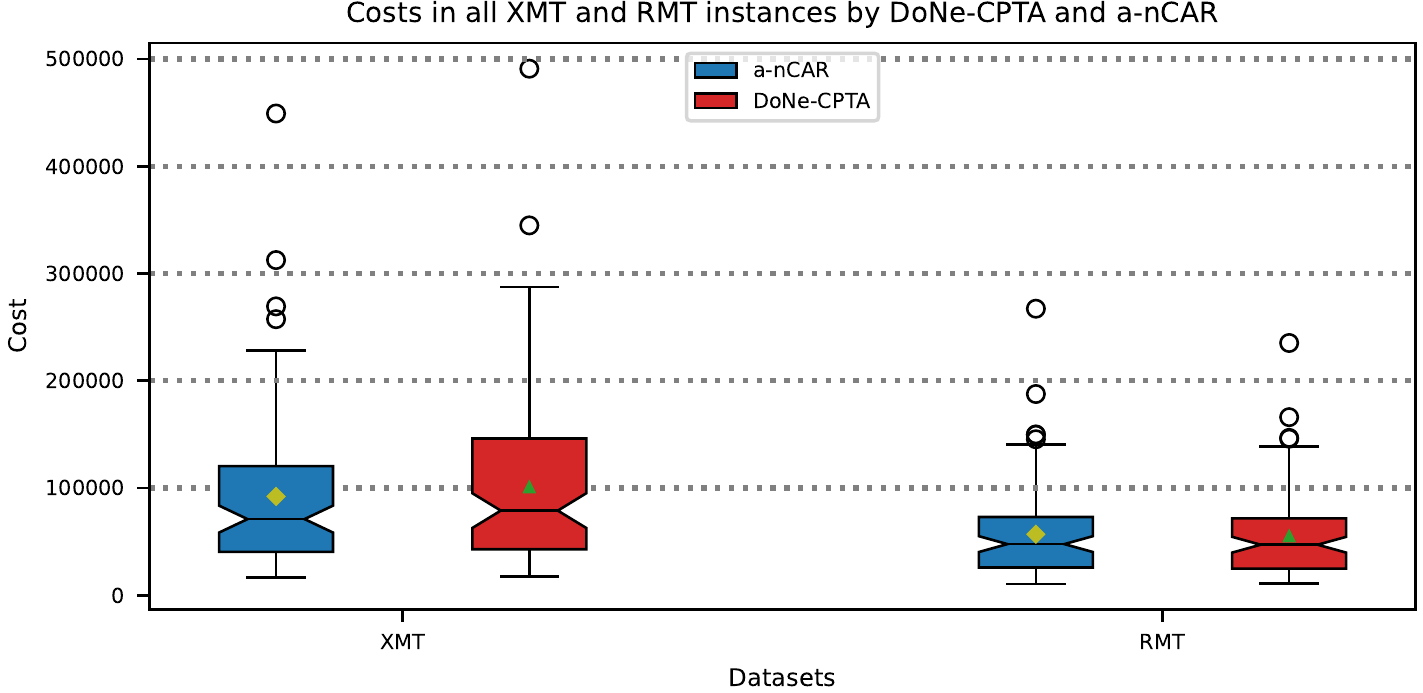}
	\caption{Overview of the cost of routes computed by DoNe-CPTA and a-nCAR when running RMT and XMT.}
	\label{fig:cost_fixedlocation_boxplot}
\end{figure*}

Figure \ref{fig:cost_fixedlocation_boxplot} presents an overview of the cost of routes in each dataset. Similar to the previous scenario, the heterogeneous fleet reduces costs even in configurations with a delivery station and robots positioned in the same location. Results indicate that DoNe-CPTA presents a similar performance compared to a-nCAR in instances with heterogeneous fleets, and on average, a slightly worse performance on homogeneous fleets. However, there was no statistical significance for these comparisons (Mann–Whitney U, $p = 0.8786$, and $p = 0.4278$, respectively). This result can be explained by the fact that these instances of the problem are much closer to the original CVPR formulation, which uses only one delivery station and fixed robots' initial locations, than the HFMDVRP-DV. Overall, this confirms the importance of having multiple delivery stations and dispersed robots along the scenario in order to improve performance on smart warehouse applications.

Since the results indicated that DoNe-CPTA and a-nCAR perform very similarly in this scenario, detailed cost results per dataset were only included in the supplementary material.

\subsubsection{Execution Time and  Number of Robots}
\label{subsubsec:evaluation2_exectime}

\begin{figure*}[ht]
	\centering
	\includegraphics[width=1\textwidth]{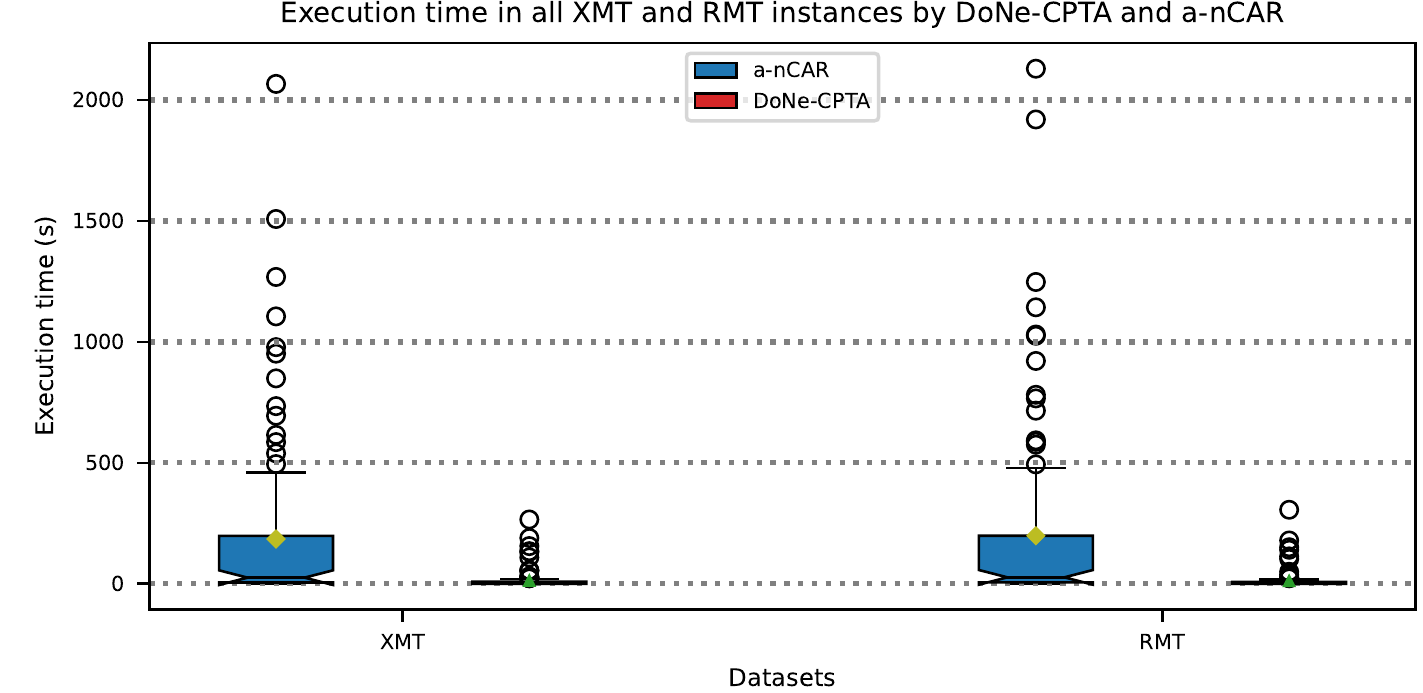}
	\caption{Overview of the execution time of DoNe-CPTA and a-nCAR when running RMT and XMT.}
	\label{fig:exectime_fixedlocationrobots_boxplot}
\end{figure*}

Execution time overview (Figure \ref{fig:exectime_fixedlocationrobots_boxplot}) shows that despite DoNe-CPTA present similar performance to a-nCAR, it still presents superior performance in route computation, i.e. execution time. The average runtime of DoNe-CPTA was $90$\% and $92$\% in XMT and RMT, respectively. Compared to the setup of dispersed robots, DoNe-CPTA was up to $5$ times slower ($15$ seconds for RMT and $16$ seconds XMT). Note that since robots are initially located at the same point, the AQ will become invalid in the initial iterations of DoNe-CPTA, increasing call to \texttt{computeDomain()} and consequently the execution time. The a-nCAR behavior was similar to the previous scenario, taking up to $200$ seconds to run most instances and over $1000$ seconds in some other cases. 



Regarding the number of robots, DoNe-CPTA is about $18$\% more efficient than a-nCAR with heterogeneous fleets. Also, our strategy employed $21$\% fewer robots in RMT compared to XMT, on average. Nevertheless, we found no evidence to assert that DoNe-CPTA employed fewer robots than a-nCAR (Mann–Whitney U, $p = 0.07911$). Also, we did not observe notable differences by nCAR in any dataset. Note that such results are not statistically significant, as we noted in Section \ref{subsubsec:evaluation1_nrobots}. All supplementary information the reader can find in the supplementary material.
\section{Conclusions}
\label{sec:conclusion}

In this paper, we present a task allocation algorithm for Multi-Robot Systems (MRS) considering a smart warehouse with automated picking \cite{Winkelhaus20211}. Our motivation is to assume that heterogeneous fleets will be constant in such warehouses. We also hypothesized that constructing a cost estimator measured as a function of all system constraints generates more accurate estimates and, consequently, more efficient task assignments. The problem was reduced to a variation of the classic Vehicle Routing Problem (VRP), called Heterogeneous Fleet Multi-Depot Vehicle Routing problem with Dispersed Vehicles (HFMDVRP-DV). The HFMDVRP-DV properties allowed us to test our hypothesis and meet the constraints in real smart warehouses. Such constraints include (i) heterogeneous robots with different load capacities and traffic speeds, (ii) multiple delivery stations, and (iii) robots dispersed across the map. Without these restrictions, the proposed problem is reduced to the Capacitated Vehicle Routing Problem (CVRP). Our main contributions:

\begin{enumerate}
	\item Present a mathematical formulation for HFMDVRP-DV since this is the first work to deal with such VRP variation, according to our most recent research.
	\item Deploy a novel cost estimator that employs all HFMDVRP-DV properties. Thus, our algorithm manages to postpone delivery task assignments because such a cost estimator penalizes costs for robots with low variable load capacity, giving way to robots performing picking tasks without visiting a delivery station.
	\item Introduce a new dataset for HFMDVRP-DV adapted from another CVRP dataset enhanced with real-world smart warehouse features.
	\item Develop DoNe-CPTA, an efficient algorithm to generate low-cost routes, with a minimum number of robots and low execution time, to solve real instances of HFMDVRP-DV.
\end{enumerate}

We also validated DoNe-CPTA against an adapted version of a state-of-the-art algorithm (a-nCAR), both running the novel HFMDVRP-DV dataset. Results showed that our strategy generates routes costing up to $33$\% less than the routes generated by a-nCAR, over $90$\% faster and using up to $18$\% fewer robots.

Our next step is to optimize domain recalculation with the expectation that DoNe-CPTA will yield better results. Our strategy currently employs a global time queue (AQ) for all robots and determines the domains' validity by comparing the value of time in the queue and whether the robot is (or is not) performing tasks. As a next step, we will determine dynamic domains using an estimate to determine the position of other robots when a particular robot achieves its task.
Our future work also includes evolving this work's contributions to deal with other real smart warehouse specifications, such as (i) order processing, (ii) delivering products from the same order at the same station, (iii) balancing in delivery station visits to avoid overloading and (iv) support tasks reallocation.

\bmhead{Acknowledgments}

The authors acknowledge the Coordenação de Aperfeiçoamento de Pessoal de Nível Superior - Brazil (CAPES) for fully funding this study.

\section*{Declarations}

\bmhead{Funding}

This study was financed in part by the Coordenação de Aperfeiçoamento de Pessoal de Nível Superior - Brazil (CAPES) - Finance Code 001, and by CAPES/PRINT - Call no. 41/2017 Senior Visiting Professor at Biomimetics and Intelligent Systems Group - BISG, University of Oulu, Finland; Finnish UAV Ecosystem (No. 338080).

\bmhead{Conflict of interest}

The authors declare that they have no conflict of interest.

\bmhead{Code and Data Availability}

The codes and datasets generated during this study, as well as the data resulting from the experiments are available at \url{https://github.com/geosoliveira/DoNe-CPTA}.

\bmhead{Authors' Contributions}

All authors contributed to the conception of this work. The code and dataset generation, literature review, and data collection were performed by George S. Oliveira. The manuscript versions up to the final version were written by George S. Oliveira, Patricia D. M. Plentz and Jônata T. Carvalho. The authors also contributed to the data analysis. Critical reviews of this work were carried out by Juha Röning, Patricia D. M. Plentz and Jônata T. Carvalho.

\bmhead{Ethics approval}

Not applicable.

\bmhead{Consent to participate}

Not applicable.

\bmhead{Consent for publication}

Not applicable.


\bibliography{references}


\end{document}